\documentclass[lettersize,journal,xllnames]{IEEEtran}
\usepackage{amsmath,amsfonts}
\usepackage{array}
\usepackage[caption=false,font=normalsize,labelfont=sf,textfont=sf]{subfig}
\usepackage{textcomp}
\usepackage{stfloats}
\usepackage{url}
\usepackage{verbatim}
\usepackage{graphicx}
\usepackage{booktabs}
\usepackage{adjustbox}
\usepackage{makecell}
\usepackage{bbding}
\newtheorem{theorem}{Theorem}[section]

\newtheorem{definition}[theorem]{Definition}
\newtheorem{assumption}[theorem]{Assumption}

\usepackage[linesnumbered,ruled]{algorithm2e}
\usepackage{multirow}
\usepackage[table]{xcolor}
\usepackage{colortbl}  
\usepackage{bm}
\definecolor{myblue}{HTML}{2878B5}
\definecolor{myred}{HTML}{C82423}
\def\BibTeX{{\rm B\kern-.05em{\sc i\kern-.025em b}\kern-.08em
    T\kern-.1667em\lower.7ex\hbox{E}\kern-.125emX}}
\usepackage{balance}
\begin{document}
\title{Interventional Imbalanced Multi-Modal Representation Learning via $\beta$-Generalization Front-Door Criterion}
\author{Yi Li$^{1,2}$, Fei Song$^{1,2}$, Changwen Zheng$^{1}$, Jiangmeng Li$^{1}$, Fuchun Sun$^{1,3}$, Hui Xiong$^{4,5}$ 
\\ $\;$
\thanks{Yi Li and Fei Song contribute equally to this work. Corresponding author: Jiangmeng Li (jiangmeng2019@iscas.ac.cn). 1. Institute of Software Chinese Academy of Sciences, Beijing, China. 2. University of Chinese Academy of Sciences, Beijing, China. 3. Tsinghua University, Beijing, China. 4. Thrust of Artificial Intelligence, the Hong Kong University of Science and Technology (Guangzhou), Guangzhou, China. 5. Department of Computer Science \& Engineering, the Hong Kong University of Science and Technology, Hong Kong SAR, China.}
}

\markboth{Journal of \LaTeX\ Class Files,~Vol.~18, No.~9, September~2020}%
{How to Use the IEEEtran \LaTeX \ Templates}

\maketitle

\begin{abstract}
Multi-modal methods establish comprehensive superiority over uni-modal methods.
However, the \textit{imbalanced} contributions of different modalities to task-dependent predictions constantly degrade the discriminative performance of canonical multi-modal methods. Based on the contribution to task-dependent predictions, modalities can be identified as \textit{predominant} and \textit{auxiliary} modalities.
Benchmark methods raise a tractable solution: augmenting the auxiliary modality with a minor contribution during training.
However, our empirical explorations challenge the fundamental idea behind such behavior, and we further conclude that benchmark approaches suffer from certain defects: insufficient theoretical interpretability and limited exploration capability of discriminative knowledge.
To this end, we revisit multi-modal representation learning from a causal perspective and build the Structural Causal Model.
Following the empirical explorations, we determine to capture the true causality between the discriminative knowledge of predominant modality and predictive label while considering the auxiliary modality.
Thus, we introduce the $\beta$-generalization front-door criterion.
Furthermore, we propose a novel network for sufficiently exploring multi-modal discriminative knowledge.
Rigorous theoretical analyses and various empirical evaluations are provided to support the effectiveness of the innate mechanism behind our proposed method.
\end{abstract}


\section{Introduction} \label{sec:intro}
A fundamental idea behind multi-modal representation learning (MML) is that the multiple modalities provide comprehensive information from different aspects, e.g., data collected from various sensors, which is inspired by the multi-sensory integration ability of humans \cite{banich2018cognitive}. Recent advances in MML \cite{hjelm2018learning, DBLP:conf/eccv/TianKI20,peng2022balanced, DBLP:journals/tkde/LiQZSRWX23} demonstrate that multiple modalities can indeed promote multi-modal models to achieve significant performance superiority over uni-modal approaches in various fields, e.g., knowledge graph \cite{DBLP:conf/nips/CaoXYHCH22,DBLP:journals/apin/LuWJHL22},
sentiment analysis \cite{kim2024msdlf,li2022clmlf,hazarika2020misa,zhu2022multimodal,huan2023unimf,wang2022cross}, and so on \cite{DBLP:journals/ai/LiZ25,yao2022multi,DBLP:journals/ai/YinZSZMZHL23,xue2024linin}.

\begin{figure*}
	\begin{center}
		{\includegraphics[width=2.0\columnwidth]{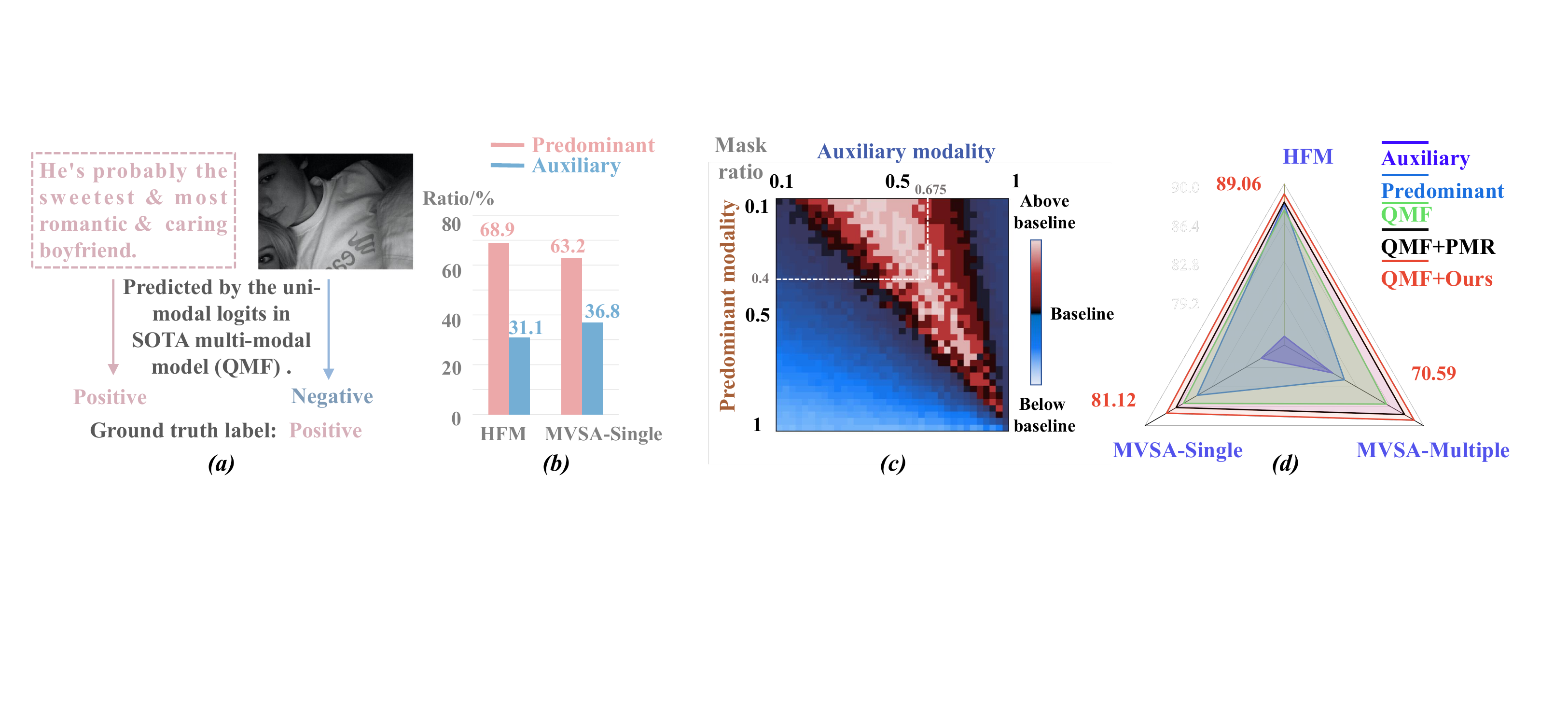}}
        \vspace{-0.3cm}
	\caption{(a): We provide an example in the MVSA-Single dataset \cite{mvsa}. Concretely, we utilize the uni-modal logits in the state-of-the-art (SOTA) multi-modal method QMF \cite{DBLP:conf/icml/ZhangWZHFZP23} to get uni-modal predictions. The emotion in the text is predicted as positive, while that of the image is predicted as negative. (b): On the two multi-modal datasets (HFM \cite{hfm} and MVSA-Single \cite{mvsa}, in which text is predominant and image is auxiliary), our statistical results indicate that, in cases where the predicted labels from the predominant and auxiliary modalities are inconsistent, the ratio that label predicted by the predominant modality is identical with the ground truth label significantly exceeds that of the auxiliary modality. (c): When evaluating the performance of QMF, we freeze all parameters of QMF and mask specific dimensions of the latent multi-modal features randomly under different ratios. We plot the performance as the heatmap, in which the lighter the color, the greater the performance boosts. (d): We depict the experimental results of various MML methods. The results demonstrate that solely utilizing the \textit{predominant} modality outperforms solely utilizing the \textit{auxiliary} modality. QMF leverages both the predominant and auxiliary modalities to achieve further performance improvement. PMR \cite{fan2023pmr} is an AMEM. \textit{QMF+PMR} augments the auxiliary modality in QMF and outperforms the plain QMF, while \textit{QMF+Ours} achieves superior performance compared to \textit{QMF+PMR}.}
		\label{fig:introduction}
	\end{center}
\end{figure*}

However, canonical MML approaches \cite{hazarika2020misa,yu2021learning,li2022clmlf} generally overlook the imbalance of different modalities, i.e., they assume that the contributions of different modalities toward the prediction of the downstream tasks are approximately balanced. Yet the theoretical and empirical basis behind such an assumption is fragile, and we conduct experimental explorations to support our statement.
As shown in Fig. (\ref{fig:introduction}a), we provide a specific illustrative example to demonstrate that the label consistency is divergent among different modalities, such that the prediction contributions of modalities are \textit{imbalanced}. As shown in Fig. (\ref{fig:introduction}b), the statistical results on the real-world datasets further prove the correctness of the above statement. Therefore, equally leveraging different modalities in MML degrades the learning of discriminative knowledge.

To address the performance degeneration of MML incurred by the imbalanced multiple modalities, the auxiliary modality enhancement methods (AMEMs) \cite{peng2022balanced,fan2023pmr,du2023uni} propose to augment the auxiliary modality during the training process, which is based on the idea that the major contribution of the predominant modality leads to the insufficient learning of the auxiliary modality. 
However, the exploration of existing benchmark methods from the dimensional perspective contradicts the behavior of the AMEMs.
As depicted in Fig. (\ref{fig:introduction}c), we mask the dimensions of multi-modal features randomly with a certain ratio, and the performance boosts are consistently observed in the upper triangular region of the heatmap.
This observation means that compared to the predominant modality, masking more dimensions of the auxiliary modality's features introduces better performance boosts, e.g., when 40\% of the predominant modality's feature and 67.5\% of the auxiliary modality's feature are masked, the performance increases, as indicated by the heatmap legend of Fig. (\ref{fig:introduction}c). This phenomenon is opposite to the AMEMs' behavior (i.e., augmenting the auxiliary modality).
Therefore, such a contradictory phenomenon demonstrates the lack of theoretical interpretability in AMEMs. Furthermore, masking the dimensions of multi-modal features leads to performance boosts, which proves the existence of noisy information detrimental to the downstream task. Thus, the discriminative knowledge explored by benchmark MML methods still has the potential to be improved.

To this end, we revisit MML from the causal perspective. Theoretically, without loss of generality, we propose a Structural Causal Model (SCM) \cite{pearl2009causality, pearl2009causal, glymour2016causal} to declare the intrinsic mechanism of introducing multiple modalities to acquire performance improvement.
From our empirical observations in Fig. (\ref{fig:introduction}b) and Fig. (\ref{fig:introduction}c), we derive an inductive conclusion: the task-dependent discriminative knowledge contained in predominant modality is superior to that of auxiliary modality, and the auxiliary modality may have certain label-inconsistent noisy information, such that \textit{naively} combining multiple modalities cannot achieve the most significant performance boosts for MML models.
Fig. (\ref{fig:introduction}d) provides sufficient evidence for the proposed inductive conclusion. From the empirical results in Fig. (\ref{fig:introduction}d), we further observe that considering the auxiliary modality, the MML model can better explore the discriminative knowledge. In this regard, according to the proposed SCM for MML, we explore the \textit{true causality} between the discriminative knowledge of the predominant modality and the ground truth label during the training process of MML models, while considering the auxiliary modality.
Thus, we introduce the \textit{$\beta$-generalization front-door criterion} and deduce the corresponding adjustment formula from the \textit{joint distribution decomposition} perspective. To better describe the intuition behind the $\beta$-generalization front-door criterion, we provide the theoretical analysis from the \textit{multi-world symbolic deduction} perspective.
Following the understanding of our theoretical findings, we implement a novel \textit{\textbf{I}nterventional imbalanced \textbf{M}ulti-\textbf{M}odal representation \textbf{L}earning} method for general MML, dubbed \textit{IMML}, which further improves the ability of MML methods in exploring discriminative knowledge from multiple modalities. Theoretically, we provide sufficient support and proof to confirm the correctness and effectiveness of IMML. In practice, the proposed IMML can function as a plug-and-play component to improve the MML performance within the imbalanced scenario. Abundant empirical results demonstrate the effectiveness of IMML consistently.
Our major contribution is four-fold:

\textbf{\romannumeral1)} We conduct empirical explorations to demonstrate the long-standing defects challenging SOTA MML methods: the insufficient theoretical interpretability and the limited ability to extract modality discriminative knowledge.

\textbf{\romannumeral2)} From the causal perspective, we theoretically propose a SCM to understand the intrinsic mechanism behind MML. To capture the true causality between the discriminative knowledge of the predominant modality and the predictive label while considering the auxiliary modality, we introduce the $\beta$-generalization front-door criterion and provide the corresponding adjustment formula with complete deduction.

\textbf{\romannumeral3)} Inspired by empirical exploration, we propose a novel network for sufficiently exploring discriminative knowledge from multiple modalities.

\textbf{\romannumeral4)} We propose IMML, which consists of the above two modules. Furthermore, this paper provides rigorous theoretical analyses and sufficient empirical evaluations to support the effectiveness of the innate mechanism behind IMML.

\section{Related Works}
\noindent \textbf{Multi-modal representation learning}. MML aims to integrate multiple modality-specific features to obtain a joint representation for downstream tasks.
Canonical MML methods treat each modality equally. For example, Self-MM \cite{yu2021learning} learns the multi-modal features by self-supervised learning, and CLMLF \cite{li2022clmlf} leverages the intrinsic attention mechanism of Transformer \cite{vaswani2017attention} to perform the multi-modal fusion.
Noting the imbalanced contributions of different modalities, AMEMs make great progress, e.g., the mutual information constraint \cite{DBLP:journals/corr/abs-2501-01240}, the gradient modulation in OGM \cite{peng2022balanced}, the prototypical method in PMR \cite{fan2023pmr} and the modality knowledge distillation in UMT \cite{du2023uni} are proposed to augment the auxiliary modality during the training process. However, AMEMs suffer from a lack of theoretical interpretability and a limited ability to explore discriminative knowledge. This paper addresses these two defects by proposing IMML, which is a new MML paradigm for the imbalanced scenario.

\noindent \textbf{Causal inference}. Because of its ability to eliminate the harmful bias of confounders and discover the causality between multiple variables \cite{pearl2009causality}, causal inference boosts the development of artificial intelligence \cite{li2021confounder,lv2022causality,mahajan2021domain}.
A widely used approach is \textit{intervention} \cite{DBLP:conf/icml/JiangCKYYW0W22, DBLP:conf/icml/QiangLZ0X22, DBLP:journals/corr/abs-2208-12681, DBLP:conf/nips/YueZS020}.
For example, based on the proposed SCM, ICL-MSR \cite{DBLP:conf/icml/QiangLZ0X22} introduces a regularization term to mitigate background disturbances through backdoor adjustment, and D\&R \cite{DBLP:journals/corr/abs-2208-12681} utilizes knowledge distillation to leverage external semantic knowledge from the causal perspective.
However, performing causal intervention via the front-door criterion has been sparsely explored \cite{xu2022neural,jeong2022finding}, and these approaches adhere to the standard constraints (three principles introduced in \cite{pearl2009causality}) to execute front-door adjustment. 
IMML is the pioneering work to introduce the $\beta$-generalization front-door criterion. Guided by this criterion, front-door adjustment can be executed under lenient constraints.

\section{Revisiting the Imbalanced MML from the Causal Perspective} \label{sec:causal}
We leverage a capital letter to represent a variable and a lowercase letter to represent its specific value.
The preliminary of causal inference is depicted in \textbf{Supplementary} \ref{sec:Preliminary of Causal Inference}. Specifically, we detail the definition of the front-door criterion \cite{pearl2009causal} and front-door adjustment \cite{pearl2009causal} in the following for ease of analysis. 
\begin{definition} \label{definition:front_door_criterion} \textit{\textbf{(Front-Door Criterion)}} \textit{A set of variables $Z$ is determined to satisfy the front-door criterion relative to an ordered pair of variables $(X, Y)$ if 
\begin{enumerate}
    \item $Z$ intercepts all directed paths from $X$ to $Y$.
    \item There is no back-door path from $X$ to $Z$.
    \item All backdoor paths from $Z$ to $Y$ are blocked by $X$.
\end{enumerate}}
\end{definition}
\begin{theorem} \label{definition:front_door_adjustment}  \emph{\textbf{(\textit{Front-Door Adjustment})}} \textit{If a set of variables $Z$ satisfy the front-door criterion relative to an ordered pair of variables $(X, Y)$, then the causal effect of $X$ on $Y$ is identifiable and is given by the following front-door adjustment formula \cite{pearl2009causal}}:
    \begin{equation}
P(Y=y \mid \text{do}(X=x)) = \sum_{z} P(z \mid x) \sum_{x'} P(y \mid x', z)P(x'). \nonumber
\end{equation}
\end{theorem}
Then we build a SCM \cite{pearl2009causal, glymour2016causal} to comprehensively understand the intrinsic mechanism behind the imbalanced MML.

\subsection{Structural Causal Model} \label{sec:scm}
\begin{figure}
\centering
\includegraphics[width=0.48\textwidth]{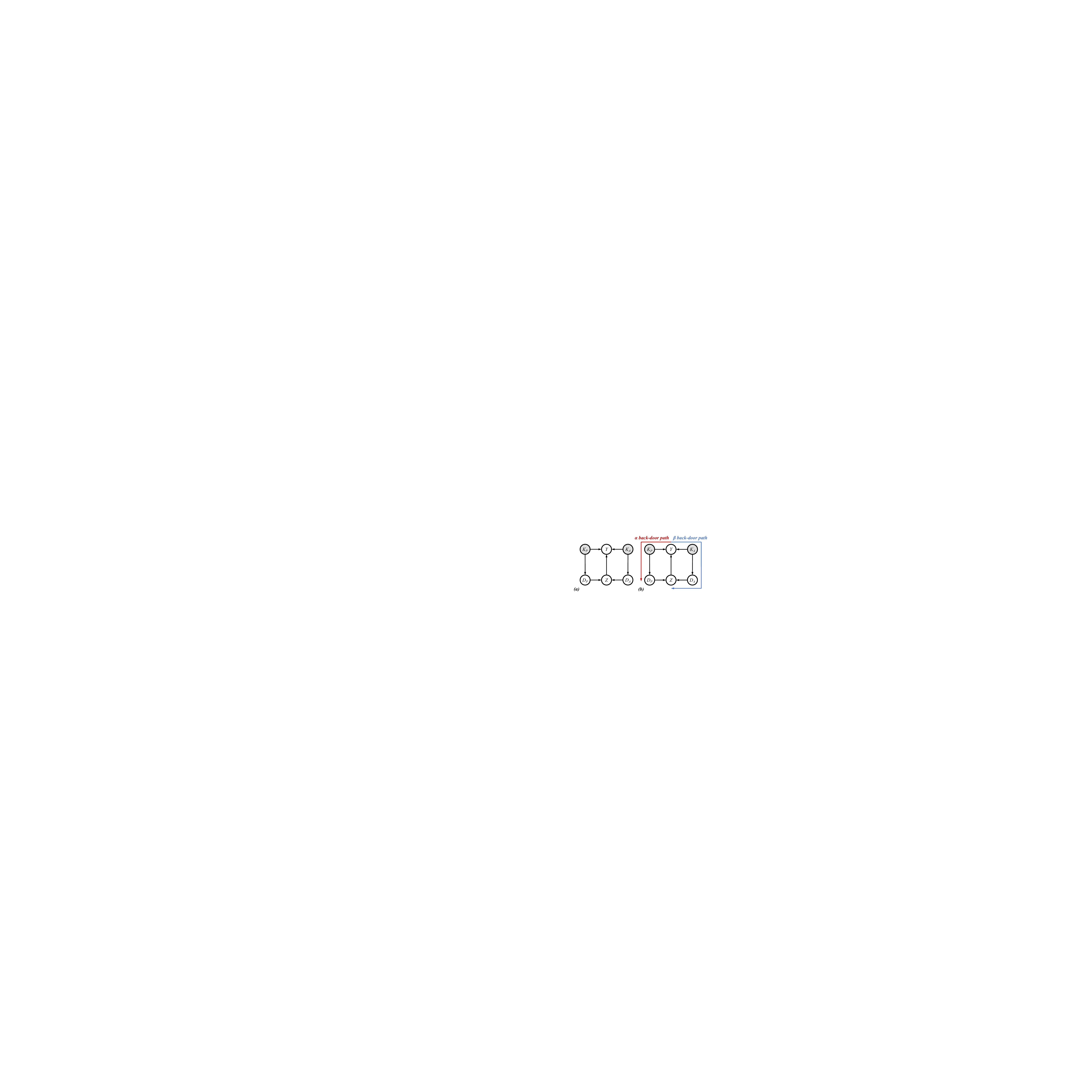}
\caption{The proposed SCM for the imbalanced MML.
a) presents the plain SCM, and b) presents the determined $\alpha$ and $\beta$ back-door paths for the proposed SCM from the perspective of the front-door criterion.}
\label{fig:scm}
\end{figure}

Following the observational exploration in \textbf{Section} \ref{sec:intro}, i.e., the existence of the predominant modality, auxiliary modality, and the confounders in the MML process, we build the SCM as demonstrated in Fig. (\ref{fig:scm}a), which holds due to the following reasons:

\noindent \textbf{\romannumeral1)} $\bm{K_P \to Y \gets K_A}$. $K_P$ and $K_A$ denote the complete knowledge 
of the predominant modality $P$ and the auxiliary modality $A$ in the MML, respectively. $Y$ denotes the corresponding predictive label. As the fundamental assumption of MML \cite{DBLP:conf/iclr/Tsai0SM21,DBLP:conf/colt/SridharanK08,DBLP:journals/corr/abs-2109-02344,DBLP:journals/tkde/LiQZSRWX23}, the knowledge of multiple modalities contains the task-dependent information, such that $Y$ is determined by $K_P$ and $K_A$ via two decoupled ways: the direct $K_P \to Y$ and $K_A \to Y$. It is worth noting that modeling the complete knowledge of $K_P$ and $K_A$ is unachievable for two reasons: 
1) the candidate inputs are sampled from the complete domain of a modality so that the available knowledge is \textit{incomplete}; 2) the knowledge modeling process is canonically performed by leveraging a non-linear neural network encoder, while according to the data processing inequality \cite{cover1999elements, DBLP:journals/tit/Willems93}, a certain inconsistency generally exists between the original knowledge of a specific modality and the corresponding modeled knowledge.
Thus $K_P$ and $K_A$ are determined as unknown in the SCM.

\noindent \textbf{\romannumeral2)} \bm{$K_P \to D_P \to Z \to Y$} and \bm{$K_A \to D_A \to Z \to Y$}. We use $D_P$ to represent the discriminative knowledge extracted by the encoder from the complete knowledge of the predominant modality. Similarly, $D_A$ signifies the discriminative knowledge gleaned from the complete knowledge of auxiliary modality. $Z$ presents the ultimate multi-modal representation formed by fusing $D_P$ and $D_A$. We reckon that $Y$ is further jointly determined by $K_P$ and $K_A$ via the mediation ways, including $K_P \to D_P \to Z \to Y$ and $K_A \to D_A \to Z \to Y$. Specifically, the reasons behind the above statement include: 1) $K_P \to D_P$ and $K_A \to D_A$: the discriminative knowledge $D_P$ and $D_A$ are extracted from $K_P$ and $K_A$ by the neural network-based encoders in MML, and the intuition behind the behavior is to model task-dependent information from multiple modalities for the prediction of $Y$; 2) $D_P \to Z \gets D_A$: the multi-modal representation $Z$ is obtained by fusing the uni-modal representations, which contains the modal-specific discriminative knowledge, i.e., $D_P$ and $D_A$; 3) $Z \to Y$: this can be implemented by the target mapping, which bridges the latent space and target space (e.g., the target mapping can be the classification layer).

\subsection{$\beta$-Generalization Front-Door Criterion}
By observing the empirical explorations in Fig. (\ref{fig:introduction}b) and Fig. (\ref{fig:introduction}c), we determine that encouraging the multi-modal representation to focus on modeling the discriminative knowledge of the predominant modality can significantly improve the performance of MML methods. To profoundly understand the phenomenon, we further analyze the experimental results in Fig. (\ref{fig:introduction}d) and find: \textbf{(\romannumeral1)} the existence of a predominant modality generally holds in MML, since learning representations solely from a certain modality consistently outperforms learning from another modality; \textbf{(\romannumeral2)} appropriately leveraging the auxiliary modality can significantly improve the model to learn task-dependent discriminative knowledge. Specifically, we conclusively determine that introducing the auxiliary modality when prompting the model to learn the causal effect between the discriminative knowledge of the predominant modality and the predictive label can improve the MML performance. By incorporating the above conclusion into the proposed SCM in Fig. (\ref{fig:scm}a), we propose sufficiently capturing the causal effect between $D_P$ and $Y$ while considering $D_A$.
In this regard, we introduce the following definitions:
\begin{definition} \label{def:alphabp}
    \textit{(\textbf{$\alpha$ Back-Door Path}) Regard $A$ and $B$ as the candidate elements within a front-door criterion scenario, and the $\alpha$ back-door path directly interferes with the estimation of the causality between $A$ and $B$.}
\end{definition}
As shown in Fig. (\ref{fig:scm}b), the back-door path $Y \gets K_P \to D_P$, interfering with the estimation of the causality between $D_P$ and $Y$, is a canonical embodiment of Definition \ref{def:alphabp}.
\begin{definition} \label{def:betabp}
    (\rm{\textbf{\textit{$\beta$ Back-Door Path}}}) \textit{Regard $A$ and $B$ as the candidate elements, and $C$ is a \textit{mediator} between $A$ and $B$ within a front-door criterion scenario. The $\beta$ back-door path \textit{indirectly} interferes with the estimation of the causality between $A$ and $B$ via the mediator $C$.}
\end{definition}
In Fig. (\ref{fig:scm}b), the back-door path $Y \gets K_A \to D_A \to Z$, which interferes with the estimation of the causality between $D_P$ and $Y$ via the mediator $Z$, exemplifies Definition \ref{def:betabp}.

The causal sub-graph, containing $K_P$, $D_P$, $Z$, and $Y$, well fits the conditions of the front-door criterion \cite{pearl2009causal, glymour2016causal}, which only includes a single back-door path between $D_P$ and $Y$, i.e., $\alpha$ back-door path, but the existence of $\beta$ back-door path violates one of the conditions of the front-door criterion, i.e., \textbf{\textit{``all backdoor paths from $Z$ to $Y$ should be blocked by $D_P$''}} \cite{pearl2009causal, glymour2016causal}. Inspired by causal applications in various fields \cite{fulcher2020robust, DBLP:conf/nips/Jeong0B22}, we propose the $\beta$-generalization front-door criterion for our proposed SCM.

\begin{definition} \label{definition:beta_front_door} \textit{(\textbf{$\beta$-generalization Front-Door Criterion}) A set of variables $Z$ is said to satisfy the $\beta$-generalization front-door criterion relative to an ordered pair of variables $(X, Y)$ if 
\begin{enumerate}
    \item $Z$ intercepts all directed paths from $X$ to $Y$.
    \item There is no back-door path from $X$ to $Z$.
\end{enumerate}}
\end{definition}

\begin{theorem} \label{definition:beta_front_door_adjustment} ({\textbf{\textit{$\beta$-generalization Front-Door Adjustment}}}) \textit{Given an observable (identifiable) variable $D_A$ on $\beta$ back-door path and a set of variables $Z$ satisfy the $\beta$-generalization front-door criterion relative to an ordered pair of variables $(D_P, Y)$, then the causal effect of $D_P$ on $Y$ is identifiable and is given by the following $\beta$-generalization front-door adjustment formula}:
    \begin{equation}
    \begin{aligned} \label{eq:main_beta_front_door_adjust}
& P(Y=y|do(D_P=d_p)) \\
& = \sum_z \sum_{d_a} \sum_{d_p'} P(y|z, d_p', d_a) P(z|d_p, d_a) P(d_a) P(d_p').
 \end{aligned}
\end{equation}
\end{theorem}

According to \cite{pearl2009causal}, Theorem \ref{definition:beta_front_door_adjustment} can be demonstrated through two distinct approaches: joint distribution decomposition and multi-world symbolic deduction, and we present the proof from these two perspectives in \textbf{Supplementary} \ref{sec:derivation of beta}.

As we can see, \textbf{\textit{$\beta$-generalization front-door criterion can be applied under fewer conditions (only requires satisfying \underline{\textbf{two out of three}} conditions in the front-door criterion), making it applicable in a wider range of scenarios.}} Accordingly, we implement our methodology by adhering to Equation (\ref{eq:main_beta_front_door_adjust}).

\section{Methodology} \label{sec:method}
\begin{figure*}
\begin{center}
{\includegraphics[width=1.6\columnwidth]{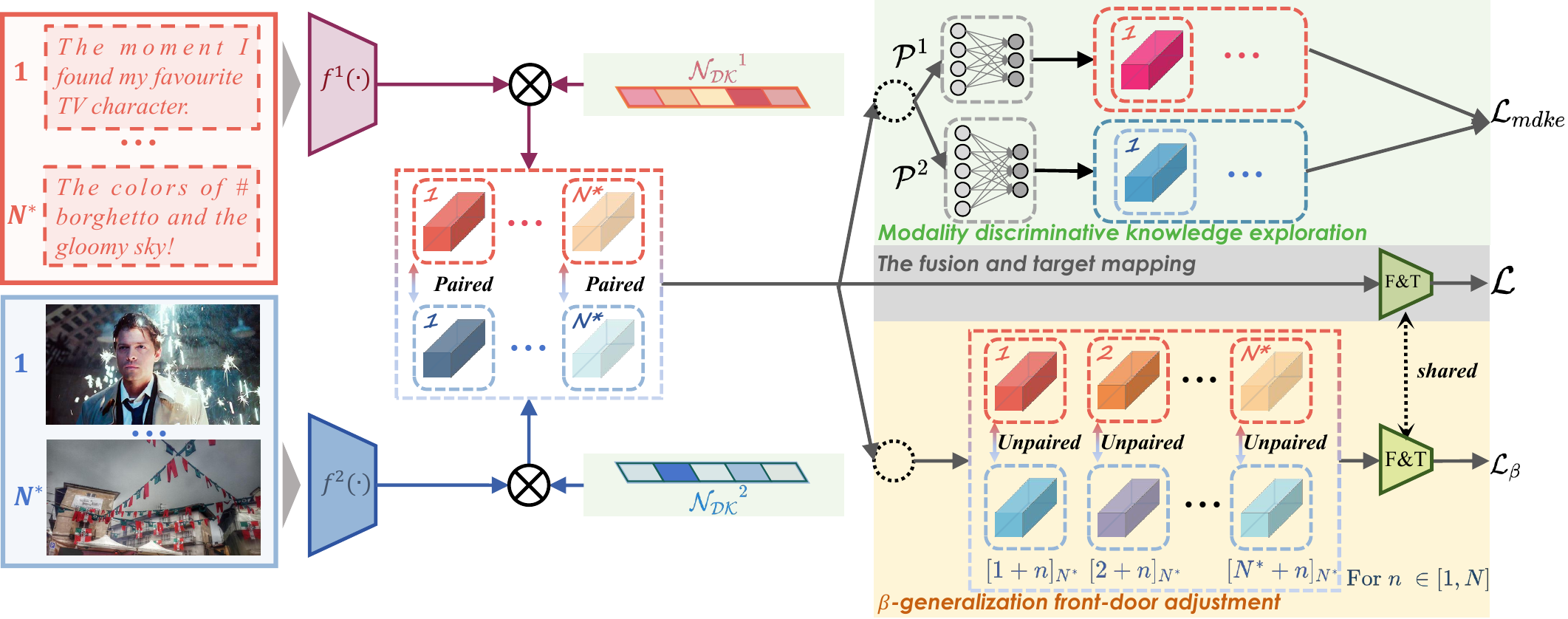}}
\vspace{-0.2cm}
\caption{We illustrate the framework of IMML with two modalities, i.e., text and image. \textit{F\&T} stands for the fusion module and target mapping module of any multi-modal model. Therefore, IMML can be treated as a plug-and-play component to boost the performance of MML within the imbalanced scenario.}
\label{fig:IMML_model}
\vspace{-0.3cm}
\end{center}
\end{figure*}

We provide an illustrative architecture of IMML in Fig. \ref{fig:IMML_model}. IMML introduces a modality discriminative knowledge exploration network to discern $D_P$ and $D_A$ from $K_P$ and $K_A$. IMML also provides a detailed functional implementation for the $\beta$-generalization front-door adjustment described above.

\subsection{Modality Discriminative Knowledge Exploration} \label{sec:mvs}

Formally, given a minibatch of multi-modal samples $\mathcal{X}=\{(x_i^m,y_i)|i\in[1,\cdots,N^*],m\in[1,\cdots,M]\}$, where $N^*$ and $M$ denote the batch size and the number of modalities, respectively. 
Specifically, the sample $x_i^m$ is first fed into the $m$-th modality-specific encoder $f^m$ (e.g., BERT \cite{BERT} for text) to obtain corresponding uni-modal feature $\boldsymbol{h}_i^m=f^m(x_i^m)$. Then we build a modality discriminative knowledge exploration network $\mathcal{N_{DK}}^m=\{\boldsymbol{\omega}^m_k|k\in[1,\cdots,D_m]\}$, where $\boldsymbol{\omega}^m_k$ is a trainable parameter and $D_m$ is the dimension of $m$-th modality's latent feature. $\mathcal{N_{DK}}^m$ assigns a weight to each dimension of the representation $\boldsymbol{h}_i^m$ by
$\hat{\boldsymbol{h}}_i^m=\boldsymbol{h}_i^m\otimes\mathcal{N_{DK}}^m$,
where $\hat{\boldsymbol{h}}_i^m \in \mathbb{R}^{D_m}$ denotes the extracted discriminative feature of the $m$-th modality and $\otimes$ is an element-wise Hadamard product function. 
Generally, the latent features of $M$ modalities have various dimensions, posing challenges in calculating the modality discriminative knowledge exploration loss. 
In this regard, we employ a projection head $\mathcal{P}^m:\mathbb{R}^{D_m} \rightarrow \mathbb{R}^D$ to obtain the dimensional-consistent latent multi-modal features $\xi_i^m = \mathcal{P}^m(\hat{\boldsymbol{h}}_i^m)$, where $\xi_i^m \in \mathbb{R}^D$. Then the loss of the modality discriminative knowledge exploration network can be formalized as
\begin{equation} \label{eq:all_cl_loss}
    \mathcal{L}_{mdke}=\sum\nolimits_{m=1}^M \mathcal{L}_{mdke}^{m,[m+1]_M},
\end{equation}
where $[x]_n = \begin{cases} 
x & \text{if } 1 \leq x \leq n \\
x \mod n & \text{if } x > n
\end{cases}$ and
{\small
\begin{equation} \label{eq:cl_loss among m and [m+1]_M1}
    \mathcal{L}_{mdke}^{m,[m+1]_M} =-\sum_{i=1}^{N^*} \log\frac{\exp\left[d\left(\boldsymbol{\xi}_i^m,\boldsymbol{\xi}_i^{[m+1]_M}\right)/\tau\right]}{\sum\limits_{i^{\prime}=1}^{N^*}\sum\limits_{m^{\prime}} \mathbb{I}_{[i\neq i^{\prime}\vee m\neq m^{\prime}]}\exp\left[d\left(\boldsymbol{\xi}_i^m,\boldsymbol{\xi}_{i^{\prime}}^{m^{\prime}}\right) / \tau\right]},
\end{equation}}
where $m^{\prime}\in \left\{m,[m+1]_M\right\}$, $d(\cdot)$ is a similarity measuring function implemented by Cosine similarity, $\mathbb{I}_{[i\neq i^{\prime}\vee m\neq m^{\prime}]}$
denotes an indicator function equalling to 1 if $i\neq i^{\prime} \; \text{or} \; m\neq m^{\prime}$
, and $\tau$ is a temperature parameter valued by following \cite{chen2020simple}. $\mathcal{L}_{mdke}$ is a variant of contrastive loss \cite{chen2020simple,wang2022chaos}, and we train $\mathcal{N_{DK}}$ using $\mathcal{L}_{mdke}$ because canonical supervised loss, e.g., cross-entropy loss \cite{de2005tutorial}, can only measure the \textit{empirical error}, whereas the introduced $\mathcal{L}_{mdke}$ can well bound the \textit{generalization error} for MML, as demonstrated by Theorem \ref{theorem:guarantee_for_generalization}. Therefore, minimizing $\mathcal{L}_{mdke}$ can improve the generalizability of IMML, thus enhancing the ability of MML models to capture discriminative knowledge from multiple modalities.

\subsection{$\beta$-Generalization Front-Door Adjustment}
This section introduces the implemented loss function for the $\beta$-generalization front-door adjustment. Without loss of generality, let $\mathcal{F}[\cdot, \cdot]$ be the arbitrary multi-modal fusion operation (e.g., $\mathcal{F}[\cdot, \cdot]$ can be concatenation, weighted summation, and so on). Let $P$ and $A$ represent the predominant and auxiliary modality, respectively. Then the dataset can be simplified as $\mathcal{X}=\{(x_i^m,y_i)|i\in[1,\cdots,N^*],m\in\{P,A\}\}$. As mentioned in \textbf{Section} \ref{sec:mvs}, we can denote the discriminative knowledge of predominant and auxiliary modalities by $\{\hat{\boldsymbol{h}}_i^m|i\in[1,\cdots,N^*],m\in[P,A]\}$. Given $D_P=
\hat{\boldsymbol{h}}_i^p$ and Equation (\ref{eq:main_beta_front_door_adjust}), $P(Y|do(D_P=\hat{\boldsymbol{h}}_i^p))$ can be rewritten as
\begin{equation} \label{eq:front_door_adj}
\begin{aligned}
&\sum_{Z} \sum_{i''=1}^{N^*} \sum_{i'=1}^{N^*}P(Y|Z,D_P=\hat{\boldsymbol{h}}_{i''}^p,D_A=\hat{\boldsymbol{h}}_{i'}^{\boldsymbol{a}}) P(D_P=\hat{\boldsymbol{h}}_{i''}^p) \\
& P(Z|D_P=\hat{\boldsymbol{h}}_i^p,D_A=\hat{\boldsymbol{h}}_{i'}^{\boldsymbol{a}})  P(D_A=\hat{\boldsymbol{h}}_{i'}^{\boldsymbol{a}}), \\
\end{aligned}
\end{equation}
which equals to
\begin{equation}
\sum_{Z}\sum_{i',i''=1}^{N^*}P(Y|Z,\hat{\boldsymbol{h}}_{i''}^p,\hat{\boldsymbol{h}}_{i'}^{\boldsymbol{a}}) P(\hat{\boldsymbol{h}}_{i''}^p)P(Z|\hat{\boldsymbol{h}}_i^p,\hat{\boldsymbol{h}}_{i'}^{\boldsymbol{a}}) P(\hat{\boldsymbol{h}}_{i'}^{\boldsymbol{a}}). \nonumber
\end{equation}

\noindent Therefore, we have transformed the summation over $D_P$ and $D_A$ into the summation over the discriminative features of the predominant and auxiliary modality. Following the computation of Equation (\ref{eq:front_door_adj}), we disclose that the calculation of $P(Z|\hat{\boldsymbol{h}}_i^p,\hat{\boldsymbol{h}}_{i'}^{\boldsymbol{a}})$ necessitates matching $\hat{\boldsymbol{h}}_i^p$ with each $\hat{\boldsymbol{h}}_{i'}^{\boldsymbol{a}}$. To avoid the excessive computation complexity, we propose to match $\hat{\boldsymbol{h}}_i^p$ with those whose indexes are close to $\hat{\boldsymbol{h}}_i^p$. Specifically, given $D_P=\hat{\boldsymbol{h}}_i^p$, we have $i'\in \{[i+1]_{N^*},\cdots,[i+N]_{N^*}\}$, where $N$ is the hyper-parameter and $[\cdot]_{N^*}$ ensures $1 \leq i' \leq N^*$.
With the intuition to fuse unpaired multi-modal features (predominant feature $\hat{\boldsymbol{h}}_i^p$ and its mismatched/unpaired features $\hat{\boldsymbol{h}}_{i'}^{\boldsymbol{a}}$), we innovatively implement
\begin{equation} \nonumber
\begin{aligned}
& z(\hat{\boldsymbol{h}}_i^p, \hat{\boldsymbol{h}}_{i'}^{\boldsymbol{a}})= z(\hat{\boldsymbol{h}}_i^p, \hat{\boldsymbol{h}}_{i'}^{a^1}, \hat{\boldsymbol{h}}_{i'}^{a^2}, \cdots, \hat{\boldsymbol{h}}_{i'}^{a^{M-1}}) \\
& =\mathcal{F}[\lambda \hat{\boldsymbol{h}}_i^p, \frac{(1-\lambda)\hat{\boldsymbol{h}}_{i'}^{a^1}}{M-1}, \frac{(1-\lambda)\hat{\boldsymbol{h}}_{i'}^{a^2}}{M-1}, \cdots, \frac{(1-\lambda)\hat{\boldsymbol{h}}_{i'}^{a^{M-1}}}{M-1}].
\end{aligned}
\end{equation}
To ensure $0 \leq \lambda \leq 1$, we sample $\lambda$ from Beta-distribution \cite{weisstein2003beta,verma2019manifold}, i.e., $\lambda \sim Beta(\alpha, \beta)$, where $\alpha$ and $\beta$ are two hyper-parameters. Meanwhile, the ground truth label of feature $z(\hat{\boldsymbol{h}}_i^p, \hat{\boldsymbol{h}}_{i'}^{\boldsymbol{a}})$ is intuitively inconsistent with the labels of $x_i^p$ or $x_{i'}^{\boldsymbol{a}}$. Thus, we redefine the ground truth label of
$z(\hat{\boldsymbol{h}}_i^p, \hat{\boldsymbol{h}}_{i'}^{\boldsymbol{a}})$ by 
$Y_z=Y_{z(\hat{\boldsymbol{h}}_i^p, \hat{\boldsymbol{h}}_{i'}^{a^1}, \hat{\boldsymbol{h}}_{i'}^{a^2}, \cdots, \hat{\boldsymbol{h}}_{i'}^{a^{M-1}})}=\lambda y(\hat{\boldsymbol{h}}_i^p) + \frac{1}{M-1} \sum_{m=1}^{M-1}(1-\lambda)y(\hat{\boldsymbol{h}}_{i'}^{a^{m}})$.
Overall, in Equation (\ref{eq:front_door_adj}), we have $P(D_A=\hat{\boldsymbol{h}}_{i'}^{\boldsymbol{a}})=\frac{1}{N}$, 
$P(D_P=\hat{\boldsymbol{h}}_{i''}^p)=\frac{1}{N^*}$, and
\begin{equation}
P(Z|\hat{\boldsymbol{h}}_i^p,\hat{\boldsymbol{h}}_{i'}^{\boldsymbol{a}})= \begin{cases}
    \frac{1}{N N^*},\quad \text{if} 
    \;\;  Z= z(\hat{\boldsymbol{h}}_i^p, \hat{\boldsymbol{h}}_{i'}^{\boldsymbol{a}}) \\
   \;\;\; 0, \quad \;\;\; \text{else}    
\end{cases}.
\end{equation}
According to the definition of $i'$, we derive the following result: $P(Y|Z,\hat{\boldsymbol{h}}_{i''}^p,\hat{\boldsymbol{h}}_{i'}^{\boldsymbol{a}})=0$ if $i^{''} \neq i$. Then, $P(Y|Z,\hat{\boldsymbol{h}}_{i''}^p,\hat{\boldsymbol{h}}_{i'}^{\boldsymbol{a}})=P(Y_z|Z=z(\hat{\boldsymbol{h}}_i^p, \hat{\boldsymbol{h}}_{i'}^{\boldsymbol{a}}))$, thus resulting in
$P(Y|do(D_P=\hat{\boldsymbol{h}}_i^p))= \frac{1}{C}\sum_{i'}P(Y_z|Z=z(\hat{\boldsymbol{h}}_i^p, \hat{\boldsymbol{h}}_{i'}^{\boldsymbol{a}}))$, where $C$ is the constant term about the probability. To capture the true causality between $D_P$ and $Y$, we determine to maximize $P(Y|do(D_P=d_p))$, i.e., minimizing the following loss function for a minibatch of multi-modal samples:
\begin{equation} \label{eq:loss_beta}
\begin{aligned}
    \mathcal{L}_{\beta} = \sum_{i=1}^{N^*} \sum\nolimits_{i'=[i+1]_{N^*}}^{[i+N]_{N^*}} l(Y_z,z(\hat{\boldsymbol{h}}_i^p, \hat{\boldsymbol{h}}_{i'}^{\boldsymbol{a}})),
\end{aligned}
\end{equation}
where $l(\cdot)$ is the loss function of the downstream task, e.g., $l(\cdot)$ can be the cross-entropy loss \cite{de2005tutorial}, mean squared error \cite{wang2009mean}, and so on. By performing the multi-task learning \cite{caruana1997multitask}, we acquire the loss function of IMML as follows:
\begin{equation}
    \mathcal{L}_{imml}  = \gamma_1 \mathcal{L}_{mdke} + \gamma_2 \mathcal{L}_{\beta} + \mathcal{L},
\end{equation}
where $\gamma_1$ and $\gamma_2$ are two hyper-parameters that control the influence of $\mathcal{L}_{mdke}$ and $\mathcal{L}_{\beta}$, respectively. $\mathcal{L}$ denotes the loss function of arbitrary benchmark MML methods, making IMML a plug-and-play component that can be generally implemented to improve various benchmarks.

\begin{table*}[t]
\caption{Classification results. \textcolor{red}{\textbf{Red}} and \textcolor{blue}{\textbf{blue}} indicate the best and second-best results, respectively. $\epsilon$ denotes the noise ratio. The dynamic model means the fusion weights in the multi-modal fusion process are functions of samples rather than constants. \XSolidBrush denotes the multi-modal fusion weights are constant, while \CheckmarkBold means the weights are the functions of samples.}
\vspace{-0.3cm}
\begin{center}
\adjustbox{max width=0.95\textwidth}{
    \begin{tabular}{c|cccccccc}
    \toprule
\multirow{3}{*}{Model}
& \multicolumn{2}{c}{Food101 \cite{food101}}
& \multicolumn{2}{c}{MVSA-Single \cite{mvsa}}
& \multicolumn{2}{c}{MVSA-Multiple \cite{mvsa}}
& \multicolumn{2}{c}{HFM \cite{hfm}} \\
       & $\epsilon=0.0$ & $\epsilon=5.0$ &  $\epsilon=0.0$ & $\epsilon=5.0$ &  $\epsilon=0.0$ & $\epsilon=5.0$ &  $\epsilon=0.0$ & $\epsilon=5.0$  \\
       \midrule
       \midrule
Bow \cite{DBLP:conf/emnlp/PenningtonSM14} (\XSolidBrush)
& 82.50 & 61.68 & 48.79 & 42.20 & 65.02 & 54.72 & 74.95 & 70.04\\
ResNet \cite{DBLP:conf/cvpr/HeZRS16} (\XSolidBrush)
& 64.62 & 34.72 & 64.12 & 49.36 & 67.08 & 60.95 & 75.82 & 73.53\\
BERT \cite{BERT} (\XSolidBrush)
& 86.46 & 67.38 & 75.61 & 69.50 & 67.59 & 64.59 & 88.09 & 82.40\\
L-f \cite{DBLP:conf/icml/ZhangWZHFZP23} (\XSolidBrush)
& 90.69 & 68.49 & 76.88 & 63.46 & 66.48 & 62.20 & 87.40 & 83.35\\
C-Bow \cite{DBLP:conf/icml/ZhangWZHFZP23} (\XSolidBrush) 
& 70.77 & 38.28 & 64.09 & 49.95 & 66.24 & 62.45 & 78.33 & 75.39\\
C-BERT \cite{DBLP:conf/icml/ZhangWZHFZP23}(\XSolidBrush) 
& 88.20 & 61.10 & 65.59 & 50.70 & 67.45 & 61.95 & 87.35 & 81.91\\
MMBT \cite{BERT} (\CheckmarkBold) 
& 91.52 & 72.32 & 78.50 & 71.99 & 67.36 & 64.22 & 87.25 & 80.92\\
TMC \cite{tmc} (\CheckmarkBold) 
& 89.86 & 73.93 & 74.88 & 66.72 & 68.65 & 64.82 & 87.31 & 83.79\\
QMF \cite{DBLP:conf/icml/ZhangWZHFZP23}  (\CheckmarkBold) 
& 92.92 & 76.03 & 78.07 & 73.85 & 69.40 & 64.81 & 87.57 & 83.90\\
\midrule
\midrule
L-f + PMR \cite{fan2023pmr}  (\XSolidBrush)
& 90.58 & 68.14 & 79.38 & 74.37 & 70.18 & 62.18 & 88.10 & 85.01\\
TMC + PMR  (\CheckmarkBold)    
& 89.72 & 73.56 & 77.84 & 70.33 & 68.26 & 64.82 & 87.51 & 83.91\\
QMF + PMR  (\CheckmarkBold)    
& 92.71 & 75.07 & 78.03 & 71.87 & 68.77 & 65.59 & 88.30 & 84.60\\
\midrule
L-f + UMT \cite{du2023uni} (\XSolidBrush) & 92.19 & 75.42 & \textcolor{blue}{\textbf{80.85}} & 72.73 & 69.47 & 65.71 & 88.22 & 85.16 \\
TMC + UMT  (\CheckmarkBold)    & 90.94 & 74.07 & 77.76 & 70.99 & 69.88 & 66.21 & 87.55 & 84.07\\
QMF + UMT (\CheckmarkBold) & \textcolor{blue}{\textbf{93.27}} & \textcolor{blue}{\textbf{76.01}} & 80.07 & 74.28 & \textcolor{blue}{\textbf{70.53}} & \textcolor{red}{\textbf{67.47}} & 88.61 & \textcolor{blue}{\textbf{84.88}}\\
\midrule
\midrule
  L-f + IMML  (\XSolidBrush) & 92.38 & 75.38 & 80.73 & \textcolor{red}{\textbf{76.88}} & 70.47 & 65.64 & \textcolor{blue}{\textbf{88.96}} & 84.41\\
 TMC + IMML  (\CheckmarkBold) & 91.30 & 74.71 & 77.65 & 67.24 & 70.23 & 66.06 & 87.46 & 84.61 \\
 QMF + IMML  (\CheckmarkBold) & \textcolor{red}{\textbf{93.46}} & \textcolor{red}{\textbf{76.31}} & \textcolor{red}{\textbf{81.12}} & \textcolor{blue}{\textbf{74.76}} & \textcolor{red}{\textbf{70.59}} & \textcolor{blue}{\textbf{66.53}} & \textcolor{red}{\textbf{89.06}} & \textcolor{red}{\textbf{85.56}} \\
    \bottomrule
    \end{tabular}}
\label{tab:main_result}
\end{center}
\vspace{-0.6cm}
\end{table*}

The training pipeline of IMML is depicted in Algorithm \ref{alg:IMML}.

\begin{algorithm}[t]
\DontPrintSemicolon
  \SetAlgoLined
  \KwIn {The sampled minibatch datasets $\mathcal{X}= \{(x_i^m,y_i)|i\in[1,\cdots,N^*],m\in[1,\cdots,M]\}$. The benchmark multi-modal $\mathcal{M}$. The hyper-parameters $\gamma_1,\gamma_2$.}
  \KwOut {The loss function of IMML $\mathcal{L}_{imml}$.}
  
  \For{$i$=1 to $N^*$}{
  Obtain uni-modal discriminative features by $ \hat{\boldsymbol{h}}_i^m=f^m(x_i^m)\otimes\mathcal{N_{DK}}^m$;

  Use $\xi_i^m=\mathcal{P}^m(\boldsymbol{h}_i^m)$ to calculate the modality discriminative knowledge loss $\mathcal{L}_{mdke}^i$  by Equation (\ref{eq:all_cl_loss}) and (\ref{eq:cl_loss among m and [m+1]_M1});

  \For{$n$=1 to N}{
  Get unpaired uni-modal features $(\hat{h}_i^m,\hat{h}_{i+n}^m)$;

Calculate $\mathcal{L}_{\beta}^i$ for $\beta$-generalization front-door adjustment by Equation (\ref{eq:loss_beta});
  }

  Calculate the loss function $\mathcal{L}^i$ of $\mathcal{M}$; 
}
\textbf{Return} $\mathcal{L}_{imml}= \sum_{i=1}^{N^*}(\gamma_1 \mathcal{L}_{mdke}^i + \gamma_2 \mathcal{L}_{\beta}^i + \mathcal{L}^i)$.
  \caption{The training pseudo code of IMML.}
  \label{alg:IMML}
\end{algorithm}

\section{Theoretical Analysis} \label{sec:theory}
We confirm that the generalization error of MML is well bounded by $\mathcal{L}_{mdke}$ with rigorous theoretical proofs. To present the connection between the generalization error and $\mathcal{L}_{mdke}$, we introduce a fundamental assumption: 
\begin{assumption} \label{assumption:label_consistency} \textbf{(Uni-modal label consistency in MML).} \textit{Suppose that the labels of paired uni-modal data are identical, i.e., $\forall$ $m_1, m_2 \in [1,\cdots,M]$, $Y(x^{m_1}_i)=Y(x^{m_2}_i)$.}
\end{assumption}
Indeed, Assumption \ref{assumption:label_consistency} is practical and can be easily achieved in real-world scenarios. For example, during the data annotation process, only the image and text pairs with consistently assigned labels are retained as data samples in the dataset \cite{mvsa}. Considering that the SOTA multi-modal classification models (MMBT \cite{mmbt}, TMC \cite{tmc}, and QMF \cite{DBLP:conf/icml/ZhangWZHFZP23}) employ a linear classification layer as target mapping and use cross-entropy loss function, without loss of generality, we derive the Theorem \ref{theorem:guarantee_for_generalization} based on the mentioned theoretical condition and the achievable Assumption \ref{assumption:label_consistency}.
\begin{theorem} \label{theorem:guarantee_for_generalization}
    \textit{\textbf{(The upper bound of generalization error)}}. Let $\mathcal{M}$ be the multi-modal model with a linear classification layer and satisfy the practical Assumption \ref{assumption:label_consistency}. Then for a $K$-class classification task, the generalization error of $\mathcal{M}$ can be bounded by $\mathcal{L}_{mdke}$:
    {\small
    \begin{equation}
    \begin{aligned}
       & GError(\mathcal{M}) \leq \sum_{m=1}^M \mathbb{E}(\phi_m) \mathbb{E}[ \mathcal{L}_{mdke}(\mathcal{N}f^m(x^m))
     + \\ & \sqrt{\mathrm{Var}(\mathcal{N}f^m(x^m)\mid y)} + \mathcal{O}( \frac{1}{\sqrt{2N^*-2}})-\log \frac{2N^*-2}{K}],
    \end{aligned}
    \end{equation}}
     where $\phi_m$ is the weight of the $m$-th modality in the multi-modal fusion, $\mathcal{N}f^m=\mathcal{{N}_{DK}}^m \circ f^m$, and $\mathrm{Var}(\mathcal{N}f^m(x^m)|y)=\mathbb{E}_{p(y)}\left[\mathbb{E}_{p(x^m|y)}\|\mathcal{N}f^m(x^m)-\mathbb{E}_{p(x^m|y)}\mathcal{N}f^m(x^m)\|^2\right]$.
\end{theorem}
The proof is provided in \textbf{Supplementary} \ref{sec:proof}. From Theorem \ref{theorem:guarantee_for_generalization}, we are inspired that by minimizing the loss $\mathcal{L}_{mdke}$, we can reduce the generalization error of the model $\mathcal{M}$, thereby ensuring the performance of $\mathcal{M}$ on unseen data samples. 

In summary, the two proposed losses are well-supported theoretically. With the guarantee of causality, minimizing $\mathcal{L}_{\beta}$ can explore the true causality between the discriminative knowledge of the predominant modality and the ground truth label while considering the auxiliary modality. According to Theorem \ref{theorem:guarantee_for_generalization}, minimizing $\mathcal{L}_{mdke}$ can enhance the generalizability of MML methods.

\section{Experiments} \label{sec:exp}
\noindent \textbf{Experimental Setup.} In this subsection, we provide the introduction of baselines, the details of datasets and implementations are deferred to \textbf{Supplementary} \ref{sec:imple} for the limited space. For comprehensive comparisons, both uni-modal models and multi-modal models are selected as our baselines. Uni-modal models include Bow \cite{DBLP:conf/emnlp/PenningtonSM14}, ResNet-152 \cite{DBLP:conf/cvpr/HeZRS16} and BERT \cite{BERT}. Multi-modal baselines contain Latefusion (L-f), ConcatBow (C-Bow), ConcatBERT (C-BERT), MMBT \cite{mmbt}, TMC \cite{tmc} and QMF \cite{DBLP:conf/icml/ZhangWZHFZP23}.
Specifically, MMBT, TMC, and QMF are \textbf{dynamic} models because the multi-modal fusion weights are the functions of samples rather than constants. For L-f and C-BERT fusion, we adopt the architecture of ResNet \cite{DBLP:conf/cvpr/HeZRS16} pretrained on ImageNet \cite{deng2009imagenet} as the backbone network for image modality and pre-trained BERT \cite{BERT} for text modality.
For C-Bow fusion, we use Bow \cite{DBLP:conf/emnlp/PenningtonSM14} to replace BERT for text modality. To demonstrate the superiority of IMML over AMEMs, we integrate two recent SOTA plug-and-play AMEMs (PMR \cite{fan2023pmr} and UMT \cite{du2023uni}) with selected MML methods (i.e., L-f, TMC and QMF) for comparison.

\begin{table}
\setlength\tabcolsep{2pt} 
\caption{The extensive link prediction results on two multi-modal knowledge graph datasets.}
    \begin{center}
     \adjustbox{max width=0.5\textwidth}{\begin{tabular}{c|cccc|cccc}
    \toprule
&  \multicolumn{4}{c|}{FB-IMG}  & \multicolumn{4}{c}{WN9-IMG} \\
Model &  MRR & H@1 & H@3 & H@10 & MRR  & H@1 & H@3 & H@10 \\ 
\midrule
TransE  & .712 & .618 & .781 & .859 & .865 & .765 & .816 & .871 \\
        DistMult &   .706 & .606 & .742 & .808 & .901 & .895 & .913 & .925 \\
        ComplEx &   .808 & .757 & .845 & .892 & .908 & .903 & .907 & .928 \\
        RotatE &   .794 & .744 & .827 & .883 & .910 & .901 & .915 & .926 \\
        \midrule
        TransAE &  .742 & .691 & .785 & .844 & .898 & .894 & .908 & .922 \\ 
        IKLR &   .755 & .698 & .794 & .857 & .901 & .900 & .912 & .928 \\ 
        TBKGE &   .812 & .764 & .850 & .902 & .912 & .904 & .914 & .931 \\ 
        MMKRL &   .827 & .783 & .857 & .906 & .913 & .905 & .917 & .932 \\ 
        OTKGE &   .843 & .799 & .876  & .916 & .923 & .911 & .930 & .947 \\
        \midrule
OTKGE+IMML & \textbf{.854} & \textbf{.812} & \textbf{.887} & \textbf{.927} & \textbf{.930} & \textbf{.916} & \textbf{.937} & \textbf{.955} \\
    \bottomrule
    \end{tabular}}
    \end{center}
\label{tab:mkg_performance}
\end{table}

\begin{table}[t]
\caption{The $p$-value in student $t$-test on four multi-modal datasets.}
 \adjustbox{max width=0.45\textwidth}{
    \begin{tabular}{c|c|c|c|c}
    \toprule[1pt]
Dataset & $\epsilon$ & \thead{L-f+IMML \\ vs L-f}  & \thead{TMC+IMML \\vs TMC} & \thead{QMF+IMML \\vs QMF}  \\
 \midrule
\multirow{2}{*}{Food101} & 0.0 & 2.84$e^{-6}$ & 2.68$e^{-6}$ & 7.02$e^{-6}$ \\
                         & 5.0 & 6.62$e^{-7}$ & 4.31$e^{-5}$ & 6.55$e^{-3}$ \\
\midrule
\multirow{2}{*}{MVSA-Single} & 0.0 & 2.38$e^{-5}$ & 6.62$e^{-6}$ & 1.21$e^{-7}$ \\
                         & 5.0 & 8.04$e^{-7}$ & 3.46$e^{-4}$ & 2.24$e^{-5}$ \\
\midrule
\multirow{2}{*}{MVSA-Multiple} & 0.0 & 3.67$e^{-5}$ & 1.29$e^{-5}$ & 1.35$e^{-5}$ \\
                         & 5.0 & 7.05$e^{-6}$ & 2.19$e^{-5}$ & 4.22$e^{-6}$ \\
\midrule
\multirow{2}{*}{HFM} & 0.0 & 2.61$e^{-5}$ & 1.97$e^{-2}$ & 3.77$e^{-6}$ \\
                         & 5.0 & 8.29$e^{-5}$ & 1.34$e^{-5}$ & 3.01$e^{-6}$ \\
\bottomrule[1pt]
\end{tabular}}
\label{tab:sig_test}
\end{table}

\begin{table}[t]
\setlength\tabcolsep{6pt} 
    \caption{The results of ablation study.}
\centering
    \adjustbox{max width=1\textwidth}{\begin{tabular}{c|cccc}
    \toprule
Model &  Food101 & M-S & M-M & HFM \\
\midrule
L-f + IMML w/o $\mathcal{L}_{mdke}$ & 91.73 & 79.96 & 69.58 & 88.45\\
L-f + IMML w/o $\mathcal{L}_{\beta}$ & 91.47 & 80.35 & 69.41 & 88.36 \\
L-f + IMML & 92.38 & 80.73 & 70.47 & 88.96\\
\midrule
TMC + IMML w/o $\mathcal{L}_{mdke}$ & 90.64 & 75.92 & 69.17 & 87.11\\
TMC + IMML w/o $\mathcal{L}_{\beta}$ & 90.86 & 75.53 & 68.59 & 87.36\\
TMC + IMML & 91.30 & 77.65 & 70.23 & 87.46\\
\midrule
QMF + IMML w/o $\mathcal{L}_{mdke}$ & 93.37 & 80.15 & 69.65 & 88.70 \\
QMF + IMML w/o $\mathcal{L}_{\beta}$ & 93.29 & 79.19 & 70.24 & 88.41\\
QMF + IMML     & 93.46 & 81.12 & 70.59 & 89.06\\
    \bottomrule
    \end{tabular}}
\label{tab:abalation_study}
\end{table}


\begin{figure*}[t]
\centering
\includegraphics[width=1\textwidth]{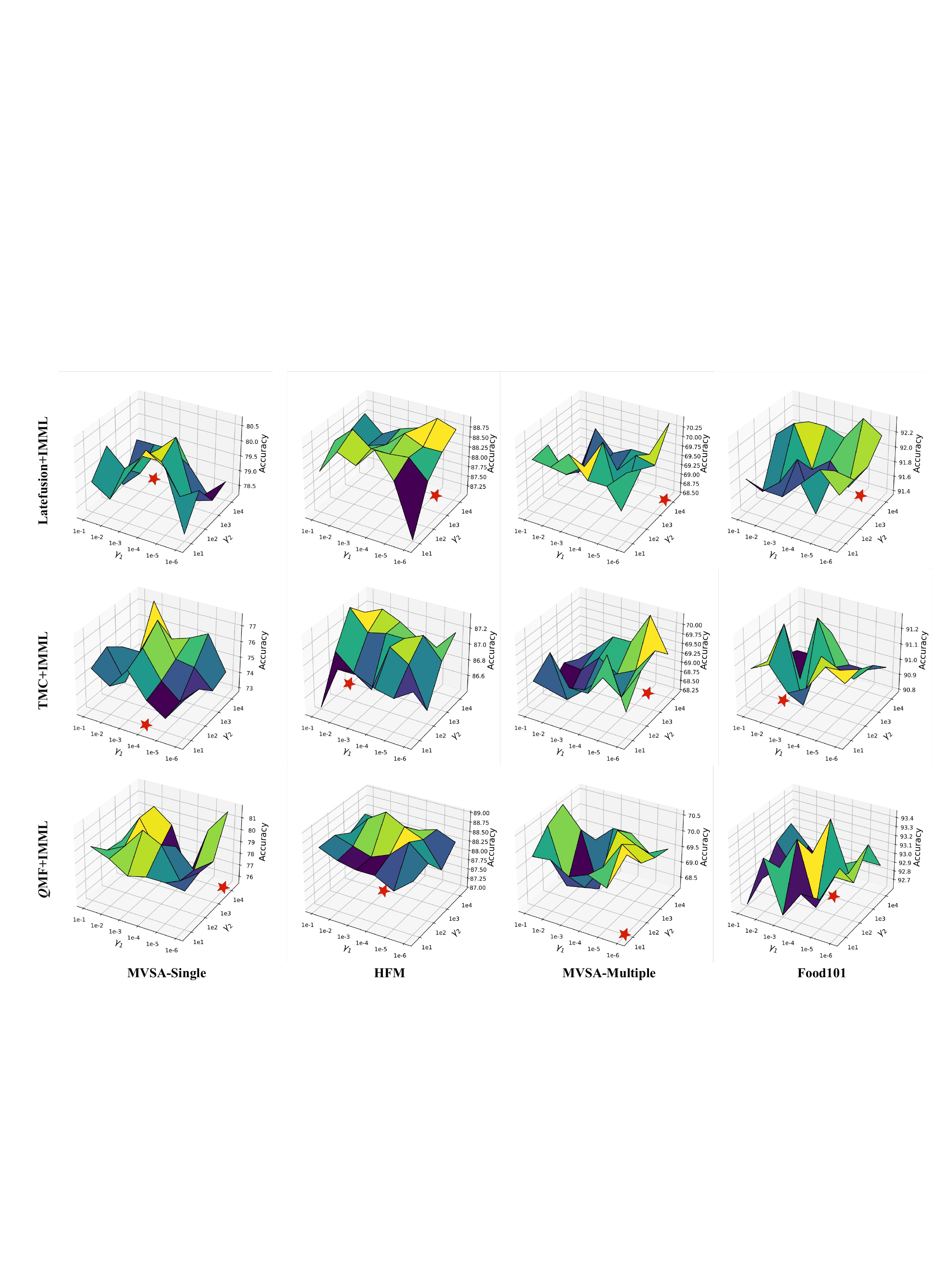}
\caption{The extended research of $\gamma_1$ and $\gamma_2$ on MVSA-Single, HFM, MVSA-Multiple and Food101.}
\label{fig:extendted_heatmap}
\vspace{-0.3cm}
\end{figure*}

\noindent \textbf{Experimental Results.}
To facilitate comprehensive comparisons, as per the baselines \cite{DBLP:conf/icml/ZhangWZHFZP23,tmc}, we introduce  Gaussian noise to the images and blank noise to the texts to assess the robustness. The overall results are depicted in TABLE \ref{tab:main_result}, and two salient observations emerge:
(\romannumeral1) Combined with IMML, benchmark MML methods exhibit significant improvements in classification accuracy, e.g., 3.05\% for QMF and 3.85\% for L-f on MVSA-Single ($\epsilon=0$), and 1.66\% for QMF on HFM ($\epsilon=5.0$).
(\romannumeral2) Compared to SOTA AMEMs, IMML achieves top-2 performance across four datasets. It's worth noting that UMT utilizes a powerful multi-modal pre-trained model (CLIP \cite{CLIP}, which is pre-trained on 400 million pairs of images and texts) to conduct knowledge distillation, thereby improving the learning of features in benchmark MML methods. Therefore, it is probably unfair to compare UMT and IMML. However, for a thorough evaluation, we still compare UMT with IMML and find that IMML outperforms UMT in 7 out of 8 comparisons.

To further demonstrate the effectiveness and generalization of IMML, we evaluate the performance of IMML on the multi-modal knowledge graph datasets WN9-IMG and FB-IMG. WN9-IMG and FB-IMG are derived from WN18 \cite{transe} and FB15K \cite{mousselly2018multimodal}, respectively. The two multi-modal knowledge graph datasets comprise the predominant structural knowledge and the auxiliary multi-modal information including text and image. Similarly, we select both uni-modal methods and multi-modal methods as our benchmark baselines, including TransE \cite{transe}, DistMult \cite{dismult}, ComplEx \cite{complex}, RotatE \cite{rotate}, IKRL \cite{ikrl}, TBKGE \cite{tbkge}, TransAE \cite{transae}, MMKRL \cite{lu2022mmkrl}, and OTKGE \cite{DBLP:conf/nips/CaoXYHCH22}. The downstream task is the link prediction and evaluation metrics are MRR, H@1, H@3, and H@10. The results in TABLE \ref{tab:mkg_performance} indicate that IMML can also enhance the performance of the SOTA method OTKGE on the link prediction task across both datasets. These experimental results consistently demonstrate the effectiveness of IMML.

\noindent \textbf{Significance Test.} 
To verify that the performance improvement is not attributed to randomness, we perform the student $t$-test \cite{mendenhall2012introduction} between the benchmark multi-modal models and the benchmark multi-modal models integrated with IMML.
$p$-value less than $0.05$ indicates that the improvement over the baseline multi-modal models is significant.
The results are shown in TABLE \ref{tab:sig_test}, confirming that the performance improvement is significant.

\noindent \textbf{Ablation Study.}
IMML consists of two vital loss functions: $\mathcal{L}_{mdke}$ and $\mathcal{L}_{\beta}$. To verify the effectiveness of each component, we conduct the ablation study, and the results are shown in TABLE \ref{tab:abalation_study}. We observe that removing any component decreases accuracy, confirming the effectiveness of both $\mathcal{L}_{mdke}$ and $\mathcal{L}_{\beta}$.
Moreover, based on the statistics in TABLE \ref{tab:abalation_study}, IMML w/o $\mathcal{L}_{mdke}$ outperforms IMML w/o $\mathcal{L}_{\beta}$ in two-thirds of cases, confirming the superiority of leveraging the $\beta$-generalization front-door adjustment for learning informative features.


\noindent \textbf{Hyper-parameters Researches on $\gamma_1,\gamma_2$.}
$\gamma_1$ and $\gamma_2$ are two hyper-parameters to control the influence of $\mathcal{L}_{mkde}$ and $\mathcal{L}_{\beta}$. $\gamma_1$ is searched in $\{1e^{-1}, 1e^{-2}, \cdots, 1e^{-6}\}$, and $\gamma_2$ is searched in $\{1e^{1}, 1e^{2}, 1e^{3}, 1e^{4}\}$. We validate these values through experimental results and depict the results in Fig. \ref{fig:extendted_heatmap}, where the light blue indicates the higher accuracy. The optimal combination of $\gamma_1$ and $\gamma_2$ varies on MML methods. For example, when integrated with IMML, QMF, TMC, and L-f achieve their best performance on the MVSA-Single dataset with $\gamma_1$ and $\gamma_2$ set to $\{1e^{-6}, 1e^4\}$, $\{1e^{-4}, 1e^1\}$, and $\{1e^{-2}, 1e^4\}$, respectively. The results illustrate that MML methods exhibit varying sensitivity to $\gamma_1$ and $\gamma_2$. Therefore, the elaborate assignment of $\gamma_1$ and $\gamma_2$ can further help IMML to learn informative features, thereby improving the performance of multi-modal models. 



\section{Conclusion} \label{sec:con}
In this paper, we conduct exploratory experiments and derive a conclusion: benchmark MML approaches lack the theoretical interpretability and the ability to capture discriminative knowledge sufficiently from multiple modalities. To better understand MML, we perform the causal analysis and determine to capture the true causality between the discriminative knowledge of predominant modality and the task-dependent label while considering the auxiliary modality. To this end, we introduce the $\beta$-generalization front-door criterion with a solid theoretical deduction. Furthermore, we propose a novel network to explore modality discriminative knowledge sufficiently. Both theoretical and experimental analyses demonstrate the effectiveness of the proposed IMML.

\section*{Acknowledgment}
This work is supported by National Natural Science Foundation of China, Grant No. 62406313, Postdoctoral Fellowship Program, Grant No. GZC20232812, China Postdoctoral Science Foundation, Grant No. 2024M753356, 2023 Special Research Assistant Grant Project of the Chinese Academy of Sciences.


\bibliographystyle{IEEEtran}
\bibliography{references}

\begin{IEEEbiographynophoto}{Yi Li}
received the B.Sc degree in the School of Fuzhou University, Fuzhou, China. He is
currently working toward the Ph.D degree in University of Chinese Academy of Sciences, Beijing, China. His research interests include in the fields of multimodal learning and causality, including vision-language representation learning and causal representation learning.
\end{IEEEbiographynophoto}

\begin{IEEEbiographynophoto}{Fei Song} received the B.Eng. degree in 2021 from Henan University, Kaifeng, China. She is currently pursuing the Ph.D. degree in software engineering at the Institute of Software, Chinese Academy of Sciences. Her research interests include multimodal prompt learning, cross-modal semantic alignment, and description-based few-shot learning.

\end{IEEEbiographynophoto}


\begin{IEEEbiographynophoto}{Changwen Zheng}
received the Ph.D. degree in Huazhong University of Science and Technology. He is currently a professor in Institute of Software, Chinese Academy of Science. His research interests include computer graph and artificial intelligence.
\end{IEEEbiographynophoto}

\begin{IEEEbiographynophoto}{Jiangmeng Li}
received the MS degree from New York University, New York, USA, in 2018, and the Ph.D. degree from University of Chinese Academy of Sciences, Beijing, China, in 2018. He is currently an assistant professor at the Institute of Software, Chinese Academy of Sciences. His research interests include multi-modal mearning, self-supervised learning, and graph learning. He has published more than 30 papers in journals and conferences such as IEEE Transactions on Knowledge and Data Engineering (TKDE), International Journal of Computer Vision (IJCV), International Conference on Machine Learning (ICML), Conference on Neural Information Processing Systems (NeurIPS), etc.
\end{IEEEbiographynophoto}

\begin{IEEEbiographynophoto}{Fuchun Sun} (Fellow, IEEE) 
received the Ph.D. degree in computer science and technology from Tsinghua University, Beijing, China, in 1997. He is currently a Professor with the Department of Computer Science and Technology and President of Academic Committee of the Department, Tsinghua University, deputy director of State Key Lab. of Intelligent Technology and Systems, Beijing, China. His research interests include artificial intelligence, intelligent control and robotics, information sensing and processing in artificial cognitive systems, etc. He was recognized as a Distinguished Young Scholar in 2006 by the Natural Science Foundation of China. He became a member of the Technical Committee on Intelligent Control of the IEEE Control System Society in 2006. He serves as Editor-in-Chief of International Journal on Cognitive Computation and Systems, and an Associate Editor for a series of international journals including the IEEE TRANSACTIONS ON COGNITIVE AND DEVELOPMENTAL SYSTEMS, the IEEE TRANSACTIONS ON FUZZY SYSTEMS, and the IEEE TRANSACTIONS ON SYSTEMS, MAN, AND CYBERNETICS: SYSTEMS.
\end{IEEEbiographynophoto}

\begin{IEEEbiographynophoto}{Hui Xiong} (Fellow, IEEE) received his Ph.D. in Computer Science from the University of Minnesota - TwinCities, USA, in 2005, the B.E. degree in Automation from the University of Science and Technology of China (USTC), Hefei, China, and the M.S. degree in Computer Science from the National University of Singapore (NUS), Singapore. He is currently a Full Professor at Rutgers, The State University of New Jersey. His general area of research is data and knowledge engineering. He received the ICDM-2011 Best Research Paper Award and the 2017 IEEE ICDM Outstanding Service Award from Rutgers, The State University of New Jersey. He has served regularly on the organization and program committees of numerous conferences, including as the Program Co-Chair for ICDM-2013, the General Co-Chair for ICDM-2015, and the Program Co-Chair for the Research Track for KDD-2018. For his outstanding contributions to data mining and mobile computing, he was elected an ACM Distinguished Scientist in 2014. He is an Associate Editor of IEEE Transactions on Knowledge and Data Engineering and ACM Transactions on Knowledge Discovery from Data.
\end{IEEEbiographynophoto}


\newpage
\onecolumn
\clearpage
\section{Preliminary of Causal Inference} \label{sec:Preliminary of Causal Inference}
Firstly, we introduce the concept of the structural causal model, i.e., SCM. 
Formally, a SCM consists of two sets of variables $U$ and $V$, and a set of
functions $f$ that assigns each variable in $V$ a value based on the values of the other variables in the model. 
A variable $X$ is a direct cause of a variable $Y$ if $X$ appears in the function that assigns $Y$’s value. $X$ is a cause of $Y$ if it is a direct cause of $Y$, or of any cause of $Y$. The variables in $U$ are called exogenous variables, meaning that they are external to the model. Exogenous variables have no ancestors and are represented as root nodes in graphs. The variables in $V$ are endogenous. Every endogenous variable in a model is a descendant of at least one exogenous variable. Exogenous variables cannot be descendants of any other variables, and in particular, cannot be descendants of an endogenous variable. If we know the value of every exogenous variable, then using the functions in $f$, we can determine with perfect certainty the value of every endogenous variable.

\begin{figure}[h]
	\begin{center}
		{\includegraphics[width=0.8\columnwidth]{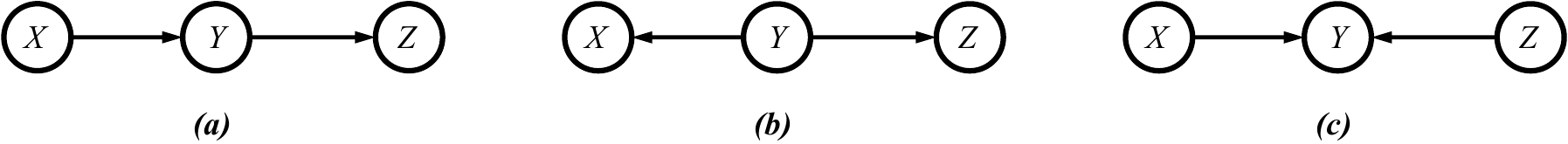}}
		\caption{The SCMs of three structures.}
		\label{fig:three_structure}
	\end{center}
\end{figure}

There are three common structures in SCM: Chain, Bifurcate, and Collision. These three structures are illustrated in Fig. (\ref{fig:three_structure}a), Fig. (\ref{fig:three_structure}b), and Fig. (\ref{fig:three_structure}c), respectively.

In the Chain structure, we have:
\begin{itemize}
\centering
    \item \textit{Z and Y are dependent.} For some $z, y$, $P(Z = z | Y = y) \neq P(Z = z)$
    \item \textit{Y and X are dependent.} For some $y, x$, $P(Y = y | X = x) \neq P(Y = y)$
    \item \textit{Z and X are likely dependent.} For some $z, x$, $P(Z = z | X = x) \neq P(Z = z)$
    \item \textit{Z and X are independent, conditional on Y.} For all $x, y, z$, $P(Z = z | X = x, Y = y) = P(Z = z | Y = y)$
\end{itemize}

In the Bifurcate structure, we have:
\begin{itemize}
\centering
    \item \textit{Z and Y are dependent.} For some $z, y$, $P(Z = z | Y = y) \neq P(Z = z)$
    \item \textit{Y and X are dependent.} For some $y, x$, $P(Y = y | X = x) \neq P(Y = y)$
    \item \textit{Z and X are likely dependent.} For some $z, x$, $P(Z = z | X = x) \neq P(Z = z)$
    \item \textit{Z and X are independent, conditional on Y.} For all $x, y, z$, $P(Z = z | X = x, Y = y) = P(Z = z | Y = y)$
\end{itemize}

In the Collision structure, we have:
\begin{itemize}
\centering
\item \textit{X and Y are dependent.} For some $x, y$, $P(X = x | Y = y) \neq P(X = x)$
    \item \textit{Z and Y are dependent.} For some $z, y$, $P(Z = z | Y = y) \neq P(Z = z)$
    \item \textit{X and Z are independent.} For all $x, z$, $P(X = x | Z = z) = P(X = x)$
    \item \textit{X and Z are dependent conditional on Y.} For some $x, y, z$, $P(X = x | Y = y, Z = z) \neq P(X = x | Y = y)$
\end{itemize}

Then we give definitions of the $d$-separation \cite{pearl2009causality}. 
\begin{definition}
\label{def:2}
\textbf{($d$-separation.)} A path $p$ is blocked by a set of nodes $Z$ if and only if:
\begin{itemize}
\item $p$ contains a chain of nodes $A \to B \to C$ or a fork $A \leftarrow B \to C$ such that the middle node $B$ is in $Z$ (i.e., $B$ is conditioned on), or

\item $p$ contains a collider $A \to B \leftarrow C$ such that the collision node $B$ is not in $Z$, and no descendant of $B$ is in $Z$. 

If $Z$ blocks every path between two nodes $X$ and $Y$,  then $X$ and $Y$ are $d$-separated, conditional on $Z$, and thus are independent conditional on $Z$.
\end{itemize}
\end{definition}

Generally, to explore the effect of $X$ on $Y$, we focus on the causal effect of $X$ on $Y$, i.e., $P(Y|do(X))$, rather than the statistical correlation between $X$ and $Y$, i.e., $P(Y|X)$. If there exists a confounder $Z$ that acts as a cause of $X,Y$ simultaneously, then 
$P(Y|do(X)) \neq P(Y|X)$.
\begin{definition}
\label{def:back_door}
\textbf{(The Backdoor Criterion.)} \textit{Given an ordered pair of variables $\left( {X,Y} \right)$ in a directed acyclic graph $G$, a set of variables $Z$ satisfies the backdoor criterion relative to $\left( {X,Y} \right)$ if no node in $Z$ is a descendant of $X$, and $Z$ blocks every path between $X$ and $Y$ that contains an arrow into $X$.}
\end{definition}

\begin{definition}
\textbf{(The Backdoor adjustment.)} \textit{If a set of variables of $Z$ satisfies the backdoor criterion for $X$ and $Y$, then the causal effect of $X$ on $Y$ is given by the formula:}
\begin{equation}
    \begin{array}{l}
P\left( {Y = y\left| {do\left( {X = x} \right)} \right.} \right)
 = \sum\limits_z {P\left( {Y = y\left| {X = x,Z = z} \right.} \right)} P\left( {Z = z} \right).
\end{array}
\end{equation}
\end{definition}

\begin{figure}[h]
	\begin{center}
		{\includegraphics[width=1\columnwidth]{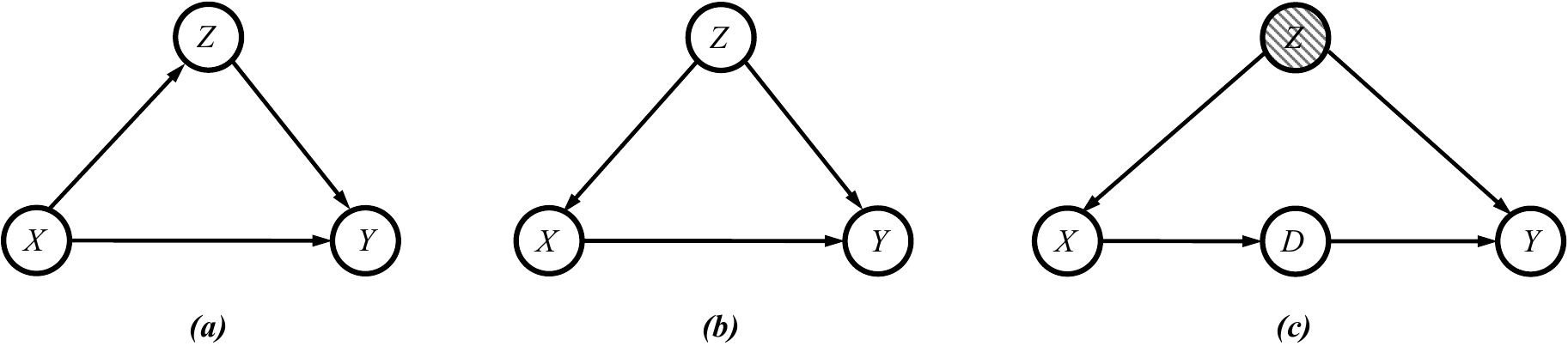}}
		\caption{The SCMs for illustration.}
		\label{fig:back_and_front}
	\end{center}
\end{figure}
For example, in Fig. (\ref{fig:back_and_front}a), no variable satisfies the backdoor criterion, thus $P(Y=y|X=x)=P(Y=y|do(X=x))$. While in Fig. (\ref{fig:back_and_front}b), we have $P\left( {Y = y\left| {do\left( {X = x} \right)} \right.} \right)
 = \sum_z {P\left( {Y = y\left| {X = x,Z = z} \right.} \right)} P\left( {Z = z} \right)$, which is not equal to $P(Y=y|X=x)$ obviously. However, as shown in Fig. (\ref{fig:back_and_front}c), when $Z$ satisfies the backdoor criterion and $Z$ is unobservable, can $P(Y|do(X))$ be identifiable or calculable? The front-door criterion is proposed to answer this question, which is depicted in \textbf{Section} \ref{sec:causal} in the main paper.

\section{Derivation of $\beta$-Generalization Front-Door Adjustment} \label{sec:derivation of beta}
In this subsection, we propose to perform the causal intervention towards the introduced SCM within the $\beta$-generalization front-door criterion scenario, thereby exploring the true causal effects between $D_P$ and $Y$, i.e., $P(Y|do(D_P=d_p))$. Formally, we present the $\beta$-generalization front-door adjustment for the proposed SCM from the perspective of the joint distribution. According to the SCM in Fig. (\ref{fig:scm}b), we formalize the corresponding joint distribution as follows:
\begin{equation} \label{eq:jointdist}
P(D_P,D_A,Z,K_A,K_P,Y) = P(K_A)P(K_P)P(D_A|K_A)P(D_P|K_P)P(Z|D_A,D_P)P(Y|Z,K_A,K_P).
\end{equation}
The $do(\cdot)$ operator removes the connections between the variable to be intervened and its parent nodes in SCM \cite{pearl2009causal}, and following our intuition, i.e., introducing $do(D_P=d_p)$, we perform the intervention on Equation (\ref{eq:jointdist}) by
\begin{equation}
        P(K_P,Z,D_A,K_A,Y|do(D_P=d_p)) = P(K_A) P(K_P)  P(D_A|K_A)P(Z|D_A,D_P=d_p)P(Y|Z,K_A,K_P), \nonumber
\end{equation}
where introducing $do(D_P=d_p)$ is equivalent to removing the term $P(D_P|K_P)$. The objective is to ascertain the causal impact of $D_P$ on $Y$. Therefore, we aggregate over the variables $Z, K_P, K_A, D_A$:
\begin{equation}
P(Y|do(D_P=d_p)) = \sum_{Z} \sum_{K_P} \sum_{K_A} \sum_{D_A} P(Z|D_P=d_p,D_A) P(D_A|K_A) P(K_A) P(K_P) P(Y|Z,K_A,K_P).
\label{eq:p(y|d_P)}
\end{equation}
As outlined in \textbf{Section} \ref{sec:scm}, $K_P, K_A$ represent the complete knowledge from the predominant and auxiliary modalities, respectively. Given that $K_P, K_A$ are unknown, it is necessary to exclude $K_P, K_A$ from Equation (\ref{eq:p(y|d_P)}). With this intuition, we introduce the following deduction:
\begin{subequations} \label{eq:derive_p(y|d_P)}
    \begin{align}
    & \sum_{Z} \sum_{K_P} \sum_{K_A} \sum_{D_A} P(Y|K_P,K_A,Z) P(D_A|K_A) P(K_A) P(K_P) P(Z|D_P=d_p,D_A) \nonumber \\
    =& \sum_{Z} \sum_{K_P} \sum_{K_A} \sum_{D_A} \sum_{d_p^{\prime} \in D_P} P(Y|K_P,K_A,Z) P(K_P|D_P=d_p^{\prime})  P(D_P=d_p^{\prime}) P(Z|D_P=d_p,D_A) \nonumber \\
    & P(D_A|K_A) P(K_A) \\
    = & \sum_{Z} \sum_{K_A} \sum_{D_A} \sum_{d_p^{\prime} \in D_P} \textcolor{red}{ \sum_{K_P} P(Y|K_P,K_A,Z,D_P=d_p^{\prime})}   \textcolor{red}{P(K_P|D_P=d_p^{\prime},Z,K_A)}  P(D_P=d_p^{\prime}) P(D_A|K_A) 
    \nonumber \\ 
    & P(K_A)  P(Z|D_P=d_p,D_A) \\ 
    = & \sum_{Z} \sum_{K_A} \sum_{D_A} \sum_{d_p^{\prime}\in D_P} \textcolor{red}{P(Y|K_A,Z,D_P=d_p^{\prime})} P(D_P=d_p^{\prime}) P(D_A|K_A) P(K_A) P(Z|D_P=d_p,D_A) \\
    = & \sum_{Z} \sum_{K_A} \sum_{D_A} \sum_{d_p^{\prime}\in D_P} P(Y|K_A,Z,D_P=d_p^{\prime}) P(D_P=d_p^{\prime}) P(K_A|D_A)  P(D_A) P(Z|D_P=d_p,D_A) \\
    = & \sum_{Z}  \sum_{D_A} \sum_{d_p^{\prime}\in D_P} \textcolor{blue}{ \sum_{K_A} P(Y|K_A,Z,D_P=d_p',D_A)} \textcolor{blue}{ P(K_A|D_A,Z,D_P=d_p')} P(Z|D_P=d_p,D_A) \nonumber \\
    & P(D_P=d_p^{\prime})  P(D_A)  \\
    = & \sum_{Z}  \sum_{D_A} \sum_{d_p^{\prime} \in D_P} \textcolor{blue}{P(Y|Z,D_P=d_p^{\prime},D_A)}  P(Z|D_P=d_p,D_A) P(D_P=d_p^{\prime})  P(D_A).
    \end{align}
\end{subequations}

Equation (\ref{eq:derive_p(y|d_P)}a) holds due to the application of the total probability equation; Equation (\ref{eq:derive_p(y|d_P)}b) holds because $Y$ is independent of $D_P$ given $K_P,K_A,Z$ and $K_P$ is independent of $Z,K_A$ given $D_P$;
Equation (\ref{eq:derive_p(y|d_P)}c) holds due to the application of the total probability equation given $D_P,Z,K_A$ (the \textcolor{red}{red} term in Equation (\ref{eq:derive_p(y|d_P)}b); Equation (\ref{eq:derive_p(y|d_P)}d) holds due to the Bayes equation (i.e., $P(K_A|D_A) = \frac{P(D_A|K_A)P(K_A)}{P(D_A)}$);
Equation (\ref{eq:derive_p(y|d_P)}e) holds because $Y$ is independent of $D_A$ given $K_A,Z,D_P$ and $K_A$ is independent of $Z,D_P$ given $D_A$;
Equation (\ref{eq:derive_p(y|d_P)}f) holds due to the application of the total probability equation given $D_P,D_A,Z$ (the \textcolor{blue}{blue} term in Equation (\ref{eq:derive_p(y|d_P)}e)).

To better demonstrate the intuition behind the behavior of $\beta$-generalization front-door adjustment, we provide the theoretical analysis from the \textit{multi-world symbolic deduction} perspective.
\begin{figure}[h]
	\begin{center}
		{\includegraphics[width=0.8\columnwidth]{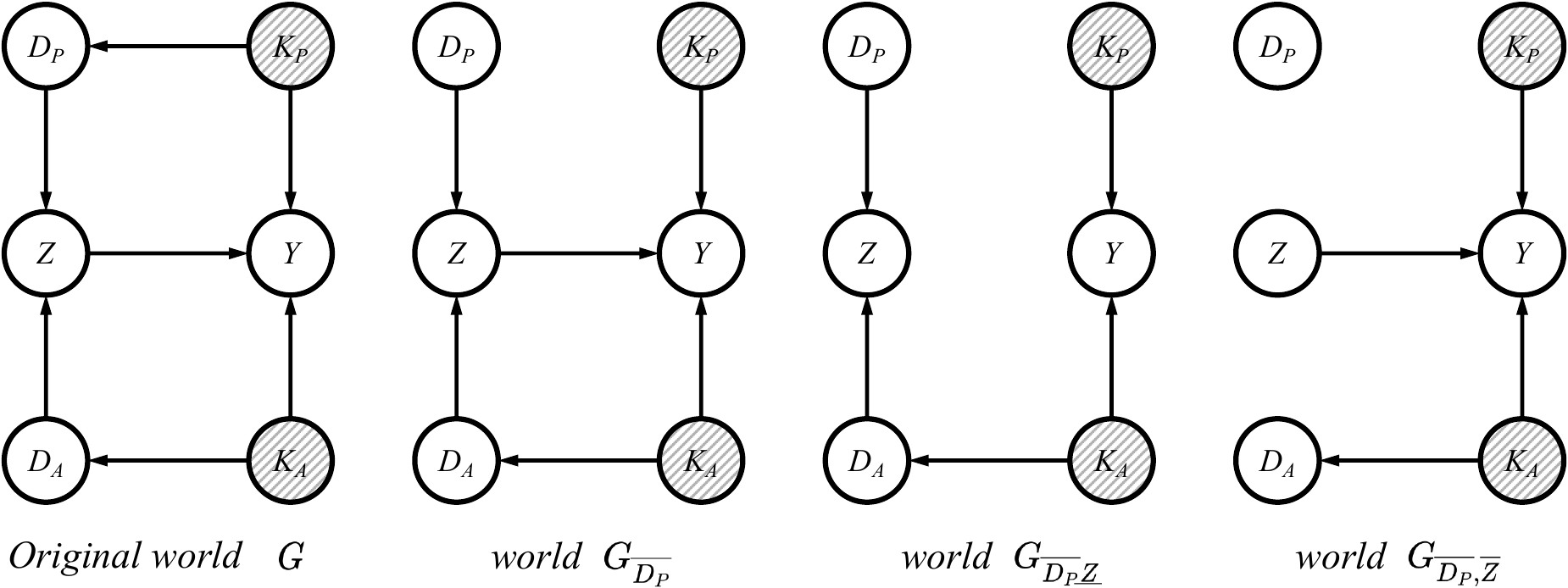}}
		\caption{The multiple worlds of original SCM.}
		\label{fig:mulple_world_scm}
	\end{center}
\end{figure}
Before we derive $P(Y|do(D_P=d_P))$ from the perspective of multi-world symbolic deduction, we introduce three rules from \cite{pearl2009causality}:
\begin{equation}
    \begin{aligned}
       & \textbf{Rule 1.} \quad \text{If} \quad (Y \perp Z | X,W)_{G_{\overline{X}}}, \quad \text{then} \quad P(Y|do(X),Z,W) = P(Y|do(X),W). \nonumber \\
       & \textbf{Rule 2.} \quad \text{If} \quad (Y \perp Z | X,W)_{G_{\overline{X}\underline{Z}}}, \quad \text{then} \quad P(Y|do(X),do(Z),W) = P(Y|do(X),Z,W). \nonumber \\
       & \textbf{Rule 3.} \quad \text{If} \quad (Y \perp Z | X,W)_{G_{\overline{X},\overline{Z(W)}}}, \quad \text{then} \quad P(Y|do(X),do(Z),W) = P(Y|do(X),W). \nonumber \\
    \end{aligned}
\end{equation}
In these three rules, $Y \perp Z$ represents that $Y$ is independent of $Z$, $G$ represents the SCM, $G_{\overline{X}}$ means removing all edges pointing to X in the SCM $G$, and $G_{\underline{Z}}$ means removing all edges pointing from Z in the SCM $G$. $Z(W)$ denotes the nodes in $G_{\overline{X}}$ that belong to $Z$ but are not ancestors of $W$.

Based on the mentioned three rules, we can derive the $P(Y|do(D_P=d_p))$ from the multi-world symbolic deduction perspective.
\begin{subequations} \label{eq:p(y|do(d_P))_multiple_world}
\begin{align}
    &P(Y|do(D_P=d_p)) = \sum_Z \sum_{D_A} P(Y|do(D_P=d_p), Z, D_A) P(Z,D_A|do(D_P=d_p)) \\
    & = \sum_Z \sum_{D_A} P(Y|do(D_P=d_p), Z, D_A) P(Z|do(D_P=d_p), D_A) P(D_A|do(D_P=d_p))\\
    & = \sum_Z \sum_{D_A} P(Y|do(D_P=d_p), Z, D_A) P(Z|do(D_P=d_p), D_A) P(D_A)\\
    & = \sum_Z \sum_{D_A} P(Y|do(D_P=d_p), Z, D_A) P(Z|D_P=d_p, D_A) P(D_A)\\
    & = \sum_Z \sum_{D_A} P(Y|do(D_P=d_p), do(Z), D_A) P(Z|D_P=d_p, D_A) P(D_A)\\
    & = \sum_Z \sum_{D_A} P(Y|do(Z), D_A) P(Z|D_P=d_p, D_A) P(D_A)\\
    & = \sum_Z \sum_{D_A} \sum_{d_p' \in D_P} P(Y|Z, D_P=d_p', D_A) P(D_P=d_p') P(Z|D_P=d_p, D_A) P(D_A)
\end{align}
\end{subequations}
Equation (\ref{eq:p(y|do(d_P))_multiple_world}a) holds due to the application of the total probability formula;
Equation (\ref{eq:p(y|do(d_P))_multiple_world}b) holds due to the conditional probability formula $P(Z,D_A)=P(Z|D_A)P(D_A)$; Equation (\ref{eq:p(y|do(d_P))_multiple_world}c) holds because $D_A$ is independent of $D_P$ in the existence of collider node $Z$; Equation (\ref{eq:p(y|do(d_P))_multiple_world}d) holds due to the invariant equation $P(Z|do(D_P), D_A)=P(Z|D_P, D_A)$ in world $G_{\overline{D_P}}$;
Equation (\ref{eq:p(y|do(d_P))_multiple_world}e) holds due to the holding of $(Y \perp Z | D_P,D_A)$ in world ${G_{\overline{D_P}\underline{Z}}}$ (Rule 2);
Equation (\ref{eq:p(y|do(d_P))_multiple_world}f) holds due to the holding of $(Y \perp D_P |Z,D_A)$ in world $G_{\overline{D_P},\overline{Z}}$ (Rule 3);
Equation (\ref{eq:p(y|do(d_P))_multiple_world}g) holds due to the normal adjustment of intervention \cite{pearl2009causal}.
To this point, we have derived the $\beta$-generalization front-door adjustment formula from the joint distribution perspective and the multi-world perspective. Furthermore, the adjustment formulas from two perspectives are consistent.

\section{Proof of Theorem \ref{theorem:guarantee_for_generalization}} \label{sec:proof}
In this section, we provide the rigorous proof for Theorem \ref{theorem:guarantee_for_generalization}. Without loss of generality, taking the $m$-th modality as an example, we denote the classification layer of the $m$-th modality as $g^m(\cdot)$. Let $\mathcal{N}f^m$ be the abbreviation of the composition function $\mathcal{{N}_{DK}}^m \circ f^m$ and $LogE = \log\mathbb{E}_{p(z_i^m)}\exp(\mathcal{N}f^m(x_i^m)^\top g^m(z_i^m))$. In practice, we can obtain the estimation of $LogE$ with $\mathcal{R}$ random samples by:
\begin{equation}
    \widetilde{LogE(\mathcal{R})} = \log \sum_{j=1}^{\mathcal{R}} \frac{1}{\mathcal{R}}\exp(\mathcal{N}f^m(x_i^m)^\top g^m(z_{i,j}^m)).
\end{equation}
Then we have:
\begin{equation} \label{eq:error}
\epsilon(\mathcal{R})=\mathbb{E}_{p(x_i^m,z^m_{i,j})}|\widetilde{LogE(\mathcal{R})}-LogE| \leq \mathcal{O}(\frac{1}{\sqrt{\mathcal{R}}}),
\end{equation}
and we provide the corresponding proof.

We have:
\begin{equation}
    \begin{aligned}
&\mathbb{E}_{p(x_i^m,z^m_{i,j})}\left[\log\frac{1}{\mathcal{R}}\sum_{j=1}^{\mathcal{R}}\exp(\mathcal{{N}} f^m(x_i^m)^\top g(z^m_{i,j}))-\log\mathbb{E}_{p(z^m_i)}\exp(\mathcal{N} f^m(x_i^m)^\top g(z^m_i))\right] \\
&\leq e\mathbb{E}_{p(x_i^m,z^m_{i,j})}\left[\frac{1}{\mathcal{R}}\sum_{j=1}^{\mathcal{R}}\exp(\mathcal{N} f^m(x_i^m)^\top g(z^m_{i,j}))-\mathbb{E}_{p(z^m_i)}\exp(\mathcal{N} f^m(x_i^m)^\top g(z^m_i))\right] =\mathcal{O}({\mathcal{R}}^{-1/2}).
\end{aligned}
\end{equation}
The first inequality holds because of Intermediate Value Theorem and $|\mathcal{{N}} f^m(x_i^m)^\top g(z^m_{i,j})| \leq 1$. The second equality holds because of Berry-Esseen Theorem, given i.i.d random variables $X_j$ with bounded support $supp(X) \in [-\alpha, \alpha]$, zero mean and bounded variance $\sigma_X^2<\alpha^2$, we have:
\begin{equation}
    \begin{aligned}
& \mathbb{E}\left[\left|\frac{1}{\mathcal{R}} \sum_{j=1}^{\mathcal{\mathcal{R}}} X_{j}\right|\right]=\frac{\sigma_{X}}{\sqrt{\mathcal{R}}} \mathbb{E}\left[\left|\frac{1}{\sqrt{\mathcal{R}} \sigma_{X}} \sum_{j=1}^{M} X_{j}\right|\right] 
= \frac{\sigma_{X}}{\sqrt{\mathcal{R}}} \int_{0}^{\frac{\alpha \sqrt{\mathcal{R}}}{\sigma_{X}}} \mathbb{P}\left[\left|\frac{1}{\sqrt{\mathcal{R}} \sigma_{X}} \sum_{j=1}^{M} X_{j}\right|>x\right] \mathrm{dx} \\
& \leq \frac{\sigma_{X}}{\sqrt{\mathcal{R}}} \int_{0}^{\frac{\alpha \sqrt{\mathcal{R}}}{\sigma_{X}}} \mathbb{P}[|\mathcal{N}(0,1)|>x]+\frac{C_{\alpha}}{\sqrt{\mathcal{R}}} \mathrm{dx}
\leq  \frac{\sigma_{X}}{\sqrt{\mathcal{R}}}\left(\frac{\alpha C_{\alpha}}{\sigma_{X}}+\int_{0}^{\infty} \mathbb{P}[|\mathcal{N}(0,1)|>x] \mathrm{dx}\right)
\leq  \frac{C_{\alpha}}{\sqrt{\mathcal{R}}}+\frac{\alpha}{\sqrt{\mathcal{R}}} \mathbb{E}[|\mathcal{N}(0,1)|]=\mathcal{O}\left(R^{-1 / 2}\right).
\end{aligned}
\end{equation}
The constant $C_{\alpha}$ depends on $\alpha$ and we set $X_j=\exp(\mathcal{N} f^m(x_i^m)^\top g(z^m_{i,j}))-\mathbb{E}_{p(z^m_i)}\exp(\mathcal{N} f^m(x_i^m)^\top g(z^m_i))$. Since $|\mathcal{{N}} f^m(x_i^m)^\top g(z^m_{i,j})| \leq 1$ and $|X_j|\leq2e$, $X_j$ has zero mean and bounded variance $(2e)^2$.

We denote the joint distribution of the positive pairs $x_i^m,x_i^{[m+1]_M}$ and the corresponding label $y_i$ by $p(x_i^m,x_i^{[m+1]_M},y_i)$. We represent the negative samples by $\{x_{i,j}^{m-}\}_{j=1}^{N_{neg}}$, where $N_{neg}=2(N^*-1)$.
Combining Equation (\ref{eq:all_cl_loss}) and Equation (\ref{eq:cl_loss among m and [m+1]_M1}), we can formalize the modality discriminative knowledge exploration loss of $m$-th modality as:

\begin{equation} \label{eq:m-th mdke loss}
\begin{aligned}
\mathcal{L}_{mdke}[\mathcal{N}f^m(x_i^m)] &= \textcolor{myblue}{\underbrace{-\mathbb{E}_{p(x_i^m,x_i^{[m+1]_M})}\mathcal{N}f^m(x_i^m)^\top\mathcal{N}f^m(x_i^{[m+1]_M})}_{\text{Term 1}}} + \textcolor{myred}{\underbrace{\mathbb{E}_{p(x_i^m)}\mathbb{E}_{p(x_{i,j}^{m-})}\log\sum\nolimits_{j=1}^{N_{neg}}\exp(\mathcal{N}f^m(x_i^m)^\top \mathcal{N}f^m(x_{i,j}^{m-}))}_{\text{Term 2}}}.
\end{aligned}
\end{equation}

Assuming the classification task has $K$ categories, we denote $\mu_y$ as the center of features from the $K$ classes. In the following, we demonstrate that the cross-entropy loss of the downstream classification task can be bounded by the proposed modality discriminative knowledge loss $\mathcal{L}_{mdke}$.

Our proof starts from Equation (\ref{eq:m-th mdke loss}).
\begin{subequations} \label{eq:term1}
    \begin{align}
&\textcolor{myblue}{\textbf{\textit{Term 1}}} = \textcolor{myblue}{-\mathbb{E}_{p(x_i^m,x_i^{[m+1]_M})}\mathcal{N}f^m(x_i^m)^\top \mathcal{N}f^m(x_i^{[m+1]_M})} \nonumber \\
& =\textcolor{myblue}{-\mathbb{E}_{p(x_i^m,x_i^{[m+1]_M},y_i)}\mathcal{N}f^m(x_i^m)^\top(\mu_{y_i}+\mathcal{N}f^m(x_i^{[m+1]_M})-\mu_{y_i})} \nonumber \\
& =\textcolor{myblue}{-\mathbb{E}_{p(x_i^m,x_i^{[m+1]_M},y_i)}\mathcal{N}f^m(x_i^m)^\top \mu_{y_i}-\mathbb{E}_{p(x_i^m,x_i^{[m+1]_M},y_i)}\mathcal{N}f^m(x_i^m)^\top(\mathcal{N}f^m(x_i^{[m+1]_M})-\mu_{y_i})} \nonumber \\
& \geq \textcolor{myblue}{-\mathbb{E}_{p(x_i^m,x_i^{[m+1]_M},y_i)}\mathcal{N}f^m(x_i^m)^\top \mu_{y_i}-\mathbb{E}_{p(x_i^m,x_i^{[m+1]_M},y_i)}\mathcal{N}f^m(x_i^m)^\top\|\mathcal{N}f^m(x_i^{[m+1]_M})-\mu_{y_i}\|} \\
& \geq \textcolor{myblue}{-\mathbb{E}_{p(x_i^m,y_i)}\mathcal{N}f^m(x_i^m)^\top\mu_{y_i}-\sqrt{\mathbb{E}_{p(x_i^m,y_i)}\|\mathcal{N}f^m(x_i^m)-\mu_{y_i}\|^2}}  \\
& \geq \textcolor{myblue}{-\mathbb{E}_{p(x_i^m,y_i)}\mathcal{N}f^m(x_i^m)^\top\mu_{y_i}-\sqrt{\operatorname{Var}(\mathcal{N}f^m(x_i^m)\mid y_i)}} \nonumber
\end{align}
\end{subequations}
Equation (\ref{eq:term1}a) holds due to $\mathcal{N}f^m(x_i^m) \in \mathbb{S}^{m-1}$ ($m$-dimensional unit sphere), which leads to: $\mathcal{N}f^m(x_i^m)^\top(\mathcal{N}f^m(x_i^{[m+1]_M})-\mu_{y_i})\leq\left(\frac{\mathcal{N}f^m(x_i^{[m+1]_M})-\mu_{y_i}}{\|\mathcal{N}f^m(x_i^{[m+1]_M})-\mu_{y_i}\|}\right)^\top(\mathcal{N}f^m(x_i^{[m+1]_M})-\mu_{y_i})=\|\mathcal{N}f^m(x_i^{[m+1]_M})-\mu_{y_i}\|$; Equation (\ref{eq:term1}b) holds due to Cauchy–Schwarz inequality and the fact that $p(x_i^m, x_i^{[m+1]_M}) = p(x_i^{[m+1]_M}, x_i^m)$ holds, where $x_i^m$ and $x_i^{[m+1]_M}$ have the same marginal distribution.

\begin{subequations} \label{eq:term2}
\begin{align}
    &\textcolor{myred}{\textbf{\textit{Term 2}}} = \textcolor{myred}{\mathbb{E}_{p(x_i^m)}\mathbb{E}_{p(x_{i,j}^{m-})}\log\sum\nolimits_{j=1}^{N_{neg}}\exp(\mathcal{N}f^m(x_i^m)^\top \mathcal{N}f^m(x_{i,j}^{m-})) \nonumber} \\  
&=\textcolor{myred}{\mathbb{E}_{p(x_i^m)}\mathbb{E}_{p(x_{i,j}^{m-})}\log\frac{1}{N_{neg}}\sum\nolimits_{j=1}^{N_{neg}}\exp(\mathcal{N}f^m(x_i^m)^\top \mathcal{N}f^m(x_{i,j}^{m-})) + \log N_{neg} \nonumber} \\ 
& \geq \textcolor{myred}{ \mathbb{E}_{p(x_i^m)}\log\frac{1}{N_{neg}}\mathbb{E}_{p(x_{i,j}^{m-})}\sum\nolimits_{j=1}^{N_{neg}}\exp(\mathcal{N}f^m(x_i^m)^\top \mathcal{N}f^m(x_{i,j}^{m-})) -\epsilon(N_{neg})+\log N_{neg}}\\ 
& =\textcolor{myred}{\mathbb{E}_{p(x_i^m)}\log\mathbb{E}_{p(x_i^{m-})}\exp(\mathcal{N}f^m(x_i^m)^\top \mathcal{N}f^m(x_i^{m-})) \nonumber -\epsilon(N_{neg})+\log N_{neg} \nonumber} \\  
&=\textcolor{myred}{\mathbb{E}_{p(x_i^m)}\log\mathbb{E}_{p(y_i^-)}\mathbb{E}_{p(x_i^{m-}|y_i^-)}\exp(\mathcal{N}f^m(x_i^m)^\top \mathcal{N}f^m(x_i^{m-})) -\epsilon(N_{neg})+\log N_{neg} \nonumber} \\ 
&\geq \textcolor{myred}{\mathbb{E}_{p(x_i^m)}\log\mathbb{E}_{p(y_i^-)}\exp(\mathbb{E}_{p(x_i^{m-}|y_i^-)}\left[\mathcal{N}f^m(x_i^m)^\top \mathcal{N}f^m(x_i^{m-})\right])-\epsilon(N_{neg})+\log N_{neg}} \\
& =\textcolor{myred}{\mathbb{E}_{p(x_i^m)}\log\mathbb{E}_{p(y_i^-)}\exp(\mathcal{N}f^m(x_i^m)^\top\mu_{y_i^-})-\epsilon(N_{neg})+\log N_{neg}} \nonumber  \\
& =\textcolor{myred}{\mathbb{E}_{p(x_i^m)}\log \frac{1}{K}\sum_{k=1}^K\exp(\mathcal{N}f^m(x_i^m)^\top\mu_k) -\epsilon(N_{neg})+\log N_{neg}} \nonumber
\end{align}
\end{subequations}
Equation (\ref{eq:term2}a) holds due to Equation (\ref{eq:error}); (\ref{eq:term2}b) holds due to the Jensen's inequality of the convex function $\exp(\cdot)$. Combining $\textbf{\textit{\textcolor{myblue}{Term 1}}}$ with $\textbf{\textit{\textcolor{myred}{Term 2}}}$, we have:
\begin{subequations} \label{eq:term1 and term2}
\begin{align}
&\text{\textbf{\textit{\textcolor{myblue}{Term 1}}}} + \text{\textbf{\textit{\textcolor{myred}{Term 2}}}} \geq
\textcolor{myblue}{-\mathbb{E}_{p(x_i^m,y_i)}\mathcal{N}f^m(x_i^m)^\top\mu_{y_i}-
\sqrt{\operatorname{Var}(\mathcal{N}f^m(x_i^m)\mid y_i)}} \nonumber \\ &+\textcolor{myred}{\mathbb{E}_{p(x_i^m)}\log \frac{1}{K} \sum_{k=1}^K\exp(\mathcal{N}f^m(x_i^m)^\top\mu_k)-\epsilon(N_{neg})+\log N_{neg}} \nonumber \\
&=\mathbb{E}_{p(x_i^m,y_i)}\big[-\mathcal{N}f^m(x_i^m)^\top\mu_{y_i}+\log\sum_{k=1}^K\exp(\mathcal{N}f^m(x_i^m)^\top\mu_k)\big] -\sqrt{\operatorname{Var}(\mathcal{N}f^m(x_i^m)\mid y_i)} \nonumber \\
&-\epsilon(N_{neg})+\log(N_{neg}/K) \nonumber \\
&=\mathcal{L}_{\mathrm{CE}}^{\mu}[\mathcal{N}f^m(x_i^m)]-\sqrt{\mathrm{Var}(\mathcal{N}f^m(x_i^m)\mid y_i)}-\epsilon(N_{neg})+\log(N_{neg}/K) \nonumber \\
& \geq \mathcal{L}_{\mathrm{CE}}[\mathcal{N}f^m(x_i^m)]-\sqrt{\mathrm{Var}(\mathcal{N}f^m(x_i^m)\mid y_i)}-\epsilon(N_{neg})+\log(N_{neg}/K).
\end{align}
\end{subequations}
As for (\ref{eq:term1 and term2}a), we have:
\begin{equation}
    \mathcal{L}_{\mathrm{CE}}^{\mu}[\mathcal{N}f^m(x_i^m)]=\mathbb{E}_{p(x_i^m,y_i)}\left[-\log\frac{\exp\left(\mathcal{N}f^m(x_i^m)^\top\mu_{y_i}\right)}{\sum_{k=1}^K\exp(\mathcal{N}f^m(x_i^m)^\top\mu_k)}\right],
\end{equation}
and thus $\mathcal{L}_{\mathrm{CE}}^{\mu}[\mathcal{N}f^m(x_i^m)] \geq min_g \; \mathcal{L}_{\mathrm{CE}}[\mathcal{N}f^m(x_i^m), g^m]$.  

Therefore, we have:
\begin{equation} \label{eq:unimodal bound}
    \mathcal{L}_{\mathrm{CE}}[\mathcal{N}f^m(x_i^m)] \leq \mathcal{L}_{mdke}[\mathcal{N}f^m(x_i^m)] + \sqrt{\mathrm{Var}(\mathcal{N}f^m(x_i^m)\mid y_i)} + \epsilon(N_{neg})-\log(N_{neg}/K).
\end{equation}
Let $\mathcal{M}$ be the multi-modal model, then:
\begin{subequations} \label{eq:final_bound}
\begin{align}
&GError(\mathcal{M}) = \mathbb{E}_{(\boldsymbol{x},y) \sim     
\mathcal{D}}\mathcal{L}_{CE}(\mathcal{N}f(\boldsymbol{x}),y)
= \mathbb{E}_{(\boldsymbol{x},y)\sim \mathcal{D}} \mathcal{L}_{CE}
(\sum_{m=1}^M \phi_m \mathcal{N}f^m(\boldsymbol{x}^m),y)  \leq \mathbb{E}_{(\boldsymbol{x},y)\sim \mathcal{D}} \sum_{m=1}^M \phi_m \mathcal{L}_{CE}(\mathcal{N}f^m(\boldsymbol{x}^m),y) \\
& = \sum_{m=1}^M \mathbb{E}_{(\boldsymbol{x},y)\sim \mathcal{D}} \phi_m \mathcal{L}_{CE}(\mathcal{N}f^m(\boldsymbol{x}^m),y) \nonumber \\
& = \sum_{m=1}^M\mathbb{E}_{(\boldsymbol{x},y)\sim \mathcal{D}}(\phi_m) \mathbb{E}_{(\boldsymbol{x},y)\sim \mathcal{D}} \mathcal{L}_{CE}[\mathcal{N}f^m(x^m),y] + Cov(\phi_m, \mathcal{L}_{CE}(\mathcal{N}f^m(x^m),y)) \nonumber \\
& \leq \sum_{m=1}^M \mathbb{E}(\phi_m) \mathbb{E}[ \mathcal{L}_{mdke}[\mathcal{N}f^m(x^m)] + \sqrt{\mathrm{Var}(\mathcal{N}f^m(x^m)\mid y)} + \epsilon(N_{neg})-\log(N_{neg}/K) \nonumber \\ & + Cov(\phi_m, \mathcal{L}_{CE}([\mathcal{N}f^m(x^m)],y))] \\
& \leq \sum_{m=1}^M \mathbb{E}(\phi_m)\mathbb{E}\left[ \mathcal{L}_{mdke}([\mathcal{N}f^m(x^m)]) + \sqrt{\mathrm{Var}([\mathcal{N}f^m(x^m)]\mid y)} + \epsilon(N_{neg})-\log(N_{neg}/K)\right].
\end{align}
\end{subequations}
Equation (\ref{eq:final_bound}a) holds due to the Jensen's inequality and the cross entropy loss function $\mathcal{L}_{CE}(\cdot)$ is convex;
Equation (\ref{eq:final_bound}b) holds due to Equation (\ref{eq:unimodal bound});
As for Equation (\ref{eq:final_bound}c), the benchmark MML methods can be divided into static and dynamic models, the fusion weights in static methods (e.g., L-f and C-BERT) are constants, thus $\phi_m$ is a constant, resulting in $Cov(\phi_m, \mathcal{L}_{CE}([\mathcal{N}f^m(x^m)],y))=0$, while the $\phi_m$ in dynamic MML methods (e.g., MMBT, TMC, and QMF)
is negatively correlated with $\mathcal{L}_{CE}([\mathcal{N}f^m(x^m)],y)$ \cite{DBLP:conf/icml/ZhangWZHFZP23,tmc}, thus $Cov(\phi_m, \mathcal{L}_{CE}([\mathcal{N}f^m(x^m)],y))\leq0$. Therefore, the Equation (\ref{eq:final_bound}c) holds.

\section{Datasets and Implementations}  \label{sec:imple}
\subsection{Datasets}
We evaluate IMML on four multi-modal classification datasets, including Food101 \cite{food101}, MVSA-Single \cite{mvsa}, MVSA-Multiple \cite{mvsa}, and HFM \cite{hfm}. The text is predominant and the image is auxiliary on these four datasets.
The evaluation metric is the accuracy. Although we conduct experiments under the condition $M=2$, it can be easily extended to cases where $M\geq 3$. Specifically, the images in Food101 are sourced from Google Image Search and accompanied by corresponding textual descriptions. MVSA-Single, MVSA-Multiple, and HFM are all collected from Twitter. TABLE \ref{datasets} presents the statistics of the four datasets, detailing the quantities of image-text pairs.

\subsection{Implementation Details}
Based on the performance of uni-modal, the text modality is chosen as our predominant modality. Since IMML is a plug-and-play component, the training setup depends on the selected MML methods. For example, the training setup of \textit{QMF+IMML} is consistent with QMF.

There are five hyper-parameters in IMML, i.e., $\alpha, \beta, \gamma_1, \gamma_2, N$. We set $\alpha=0.1, \beta=0.1$, thus $\lambda \sim Beta(0.1, 0.1)$. $\gamma_1$ and $\gamma_2$ control the influence of $\mathcal{L}_{mdke}$ and $\mathcal{L}_{\beta}$, and the range of $i'$ is determined by $N$. In practice, we search $\gamma_1$ in $\{1e^{-1},1e^{-2},1e^{-3},1e^{-4},1e^{-5},1e^{-6}\}$ and $\gamma_2$ in $\{1e^{1},1e^{2},1e^{3},1e^{4}\}$, and we set $N=2$. All experiments are conducted on four A100 GPUs.

\begin{table}[h]
\caption{Details of four datasets.}
\begin{center} 
\begin{tabular}{c|c|c|c|c}
  \toprule
  datasets &  train & test & val & total \\
  \midrule
  Food101 & 21695 & 60601 & 5000 & 87296 \\
  MVSA-single & 3611 & 450 & 450 & 4511\\
  MVSA-multiple & 13624 & 1700 & 1700 & 17024\\
  HFM & 19816 & 2410 & 2409 & 24635\\
  \bottomrule
\end{tabular}
\label{datasets}
\end{center}
\end{table}

\end{document}